\documentclass[runningheads]{llncs}
\usepackage{hyperref}
\usepackage{url}
\usepackage{float}
\usepackage{graphicx}
\usepackage[tableposition=top, labelfont=bf, labelsep=period]{caption}
\usepackage{amsmath,amsfonts,bm}

\def\eqref#1{equation~\ref{#1}}

\def\1{\bm{1}}

\DeclareMathAlphabet{\mathsfit}{\encodingdefault}{\sfdefault}{m}{sl}
\SetMathAlphabet{\mathsfit}{bold}{\encodingdefault}{\sfdefault}{bx}{n}

\DeclareMathOperator*{\argmax}{arg\,max}
\DeclareMathOperator*{\argmin}{arg\,min}

\usepackage{tikz} 
\usepackage{epsfig}
\usepackage{adjustbox}
\usepackage{algorithm}
\usepackage{algpseudocode}
\usepackage{tabularx}
\usepackage{multirow}
\usepackage{cite}

\newcommand*\diff{\mathop{}\!\mathrm{d}}

\newcommand{\citet}[1]{\cite{#1}}
\newcommand{\repeatthanks}{\textsuperscript{\thefootnote}}

\begin{document}
\title{Towards Efficient MCMC Sampling in Bayesian Neural Networks by Exploiting Symmetry}
\titlerunning{Towards Efficient MCMC Sampling in Bayesian Neural Networks}
%
\author{Jonas Gregor Wiese\thanks{Equal contribution.} \inst{1} \and Lisa Wimmer\repeatthanks\inst{2,3} \and  Theodore Papamarkou\inst{4} \and \\Bernd Bischl\inst{2,3} \and  Stephan G\"unnemann\inst{1,3} \and  David R\"ugamer\inst{2,3}}
\authorrunning{J. G. Wiese et al.}

\institute{
Technical University of Munich
\and Department of Statistics, LMU Munich
\and Munich Center for Machine Learning
\and Department of Mathematics, The University of Manchester}

\maketitle              
\begin{abstract} 
    Bayesian inference in deep neural networks is challenging due to the high-dimensional, strongly multi-modal parameter posterior density landscape.
    Markov chain Monte Carlo approaches asymptotically recover the true posterior but are considered prohibitively expensive for large modern architectures.
    Local methods, which have emerged as a popular alternative, focus on specific parameter regions that can be approximated by functions with tractable integrals.
    While these often yield satisfactory empirical results, they fail, by definition, to account for the multi-modality of the parameter posterior.
    In this work, we argue that the dilemma between exact-but-unaffordable and cheap-but-inexact approaches can be mitigated by exploiting symmetries in the posterior landscape.
    Such symmetries, induced by neuron interchangeability and certain activation functions,  manifest in different parameter values leading to the same functional output value.
    We show theoretically that the posterior predictive density in Bayesian neural networks can be restricted to a symmetry-free parameter reference set.
    By further deriving an upper bound on the number of Monte Carlo chains required to capture the functional diversity, we propose a straightforward approach for feasible Bayesian inference.
    Our experiments suggest that efficient sampling is indeed possible, opening up a promising path to accurate uncertainty quantification in deep learning.
    
\end{abstract}

\section{Introduction}

Despite big data being the dominant paradigm in deep learning, the lack of infinitely many observations makes uncertainty quantification (UQ) an important problem in the field.
A key component of UQ in Bayesian learning is the parameter posterior density that assigns a posterior probability to each parameter value\footnote{
We assume the likelihood to be parameterized by a single parameter vector. 
In the case of neural networks (NNs), the parameter contains all weights and biases.
} \cite{hullermeier_aleatoric_2021}.
Between the extreme cases of all posterior probability mass concentrating on a single value, indicating complete certainty about the model parameters, and being distributed uniformly over all possible values in a reflection of total ignorance, the shape of the parameter posterior density is central to the quantification of predictive uncertainty.
However, the parameter posterior for Bayesian neural networks (BNNs) is typically highly multi-modal and rarely available in closed form.
The classical Markov chain Monte Carlo (MCMC) approach asymptotically recovers the true posterior but is considered prohibitively expensive for BNNs, as the large number of posterior modes prevents a reasonable mixing of chains \cite{izmailov_what_2021}.
Popular approximation techniques, such as Laplace approximation (LA; \cite{mackay_1992_BayesianInterpolation, daxberger_2021_LaplaceReduxEffortless}) or deep ensembles (DE; \cite{lakshminarayanan_simple_2017}), therefore focus on local regions of the posterior landscape.
While these methods are faster than traditional MCMC and perform well in many applications, they systematically omit regions of the parameter space that might be decisive for meaningful UQ \cite{izmailov_what_2021} (also shown in Section~\ref{subsec:results_symmetry_removal}).

In this work, we challenge the presumed infeasibility of MCMC for NNs and propose to exploit the -- in this context, rarely considered -- unidentifiability property of NNs, i.e., the existence of two or more equivalent parameter values that describe the same input-output mapping.
We refer to these equivalent values as \textit{equioutput parameter states}. 
Equioutput parameter states emerge from certain activation functions \cite{kurkova_functionally_1994, chen_geometry_1993, petzka_notes_2020}, as well as the free permutability of neuron parameters in hidden layers \cite{hecht-nielsen_algebraic_1990}, and can be transformed into one another.

The functional redundancy arising from this phenomenon grows rapidly with the depth and width of a network (cf.~Figure~\ref{fig:redundancy}) and induces symmetries in the posterior density.
For exact inference (up to a Monte Carlo error), we need to incorporate all non-equioutput parameter states that lead to distinct input-output mappings. Considering only these \emph{functionally diverse} mappings means, in turn, that our effective parameter space makes up a comparatively small fraction of the network's original parameter space.
Since their numerous equioutput counterparts do not contribute any new information to predictive uncertainty, we need much fewer MCMC samples when approximating the posterior predictive density (PPD) via Monte Carlo integration.
By explicitly removing symmetries from samples \textit{post-hoc}, we can even expose the functionally relevant part of the posterior and provide an opportunity for interpretation and analytical approximation in the reduced effective parameter space.

\subsubsection{Our Contributions.}
We analyze the role of posterior space redundancies in quantifying BNN uncertainty, making the following contributions: 
1) We show that the full PPD can be obtained from a substantially smaller reference set containing uniquely identified parameter states in function space.
2) We propose an estimation procedure for the number of Monte Carlo chains required to discover functionally diverse modes, providing a practical guideline for sampling from the parameter space of multi-layer perceptrons (MLPs).
3) We supply experimental evidence that our approach yields superior predictive performance compared to standard MCMC and local approximation methods.
4) Lastly, we demonstrate the posterior interpretability and analytic approximation that can be obtained from explicitly removing symmetries \textit{post-hoc}, for which we propose an algorithmic proof-of-concept.

\section{Related Work} \label{sec:related_work}
In this section, we review the literature on the existence, removal, and utilization of symmetries in MLPs, as well as the relation between such symmetries and statistical inference.

\subsubsection{Existence of Parameter State Symmetries.}
Non-unique network parameter states have been considered in the literature before.
\cite{hecht-nielsen_algebraic_1990} were among the first to note that equioutput states induce symmetries in the parameter space of MLPs.
Focusing, within the general linear group of the parameter space, on the subgroup of transformations  that leave the input-output mapping unchanged, they derived equivalence classes of equioutput parameter states and showed that, for every MLP, there exists a minimal and complete set of representatives covering all functionally different parameter states.
\cite{sussmann_uniqueness_1992} and \citet{kurkova_functionally_1994} continued along this line of work to study single-hidden-layer MLPs with specific activation functions, advancing from tanh to more general self-affine activations. 
An extension to MLPs of arbitrary depth was studied by \cite{chen_geometry_1993} in the context of tanh activations.
More recently, \cite{petzka_notes_2020} characterized equioutput parameter states for ReLU activations, again focusing on the case of a single hidden layer,
and \cite{agrawal2022a} classified all $\mathbb{G}$-invariant single-hidden-layer MLPs with ReLU activation for any finite orthogonal group $\mathbb{G}$.
Lastly, \citet{VLACIC2021107485} generalized much of the above in a framework addressing the identifiability of affine symmetries in arbitrary architectures. 

\subsubsection{Symmetry Removal.}
Symmetries in the parameter posterior density of Bayes\-ian models can produce adverse effects that have been addressed in several research areas of statistics and machine learning.
A prominent example is \emph{label switching} in finite mixture models, where the permutability of label assignments to the mixture components induces symmetries similar to those in BNNs.
To make mixture models identifiable, \citet{bardenet_adaptive_2012} introduced an adaptive Metropolis algorithm with online relabeling, effectively removing permutation symmetries by optimizing over discrete sets of permutation matrices.
Such exhaustive-search approaches, however, scale poorly to modern NNs with many parameters, as the amount of equioutput states rises exponentially with the number of parameters.

In BNNs, symmetries have been known to slow down MCMC convergence to the stationary parameter posterior density due to budget wasted on visiting symmetric modes \cite{nalisnick_priors_2018, papamarkou_challenges_2021}.
\cite{izmailov_what_2021}, reporting results from extensive and large-scale experiments, indeed find that MCMC chains tend to mix better in function space than in parameter space.
Consequently, reducing the effect of symmetries by imposing parameter constraints and defining anchoring points for subsets of the latent variables has been shown to improve mixing \citet{sen_bayesian_2020}.
A proposal for constrained sampling can be found, for example, in \cite{pourzanjani_improving_2017}, with application to ReLU-activated MLPs.

\subsubsection{Utilizing Symmetries.}
Symmetries in the parameter posterior density are not, however, necessarily a nuisance. 
Quite on the contrary, they can be useful to enhance generalization and make inference affordable.
An increasing body of work has been exploring the use of symmetry removal in the context of \emph{mode connectivity}, an approach to find more robust NN solutions by retrieving connected areas of near-constant loss rather than isolated local optima \cite{draxler_essentially_2018, garipov_loss_2018}.
Focusing on equioutput permutations of hidden-layer neurons, \citet{tatro_optimizing_nodate} and \citet{ainsworth2023git}, among others, propose to align the layer-wise embeddings of multiple networks and thus improve upon the performance of individual models.
Following a similar idea, \citet{pittorino_2022_DeepNetworksToroidsb} apply a \textit{post-hoc} standardization on parameter vectors in ReLU-activated NNs that draws from the notion of equioutput equivalence classes.

In the field of Bayesian deep learning, the idea of utilizing -- exact or approximate -- parameter symmetries, represented by permutation groups, has led to the development of \emph {lifted MCMC}~\cite{niepert2013, broeck2021}.
Orbital Markov chains have been introduced to leverage parameter symmetries in order to reduce mixing times~\cite{niepert2012b, broeck2021}.
Lifted MCMC has been considered mainly in the context of probabilistic graphical models.
There is scope to harness lifted MCMC in the context of MLPs since these can be cast as graphical models \cite{kipf_2017_SemiSupervisedClassificationGraph, VLACIC2021107485}. \\


We believe that equioutput symmetries have the potential to facilitate MCMC inference, despite the apparent complexity they introduce in the parameter posterior density.
From the insight that a vast part of the sampling space is made up of symmetric copies of some minimal search set, we conclude that running multiple short MCMC chains in parallel, each of which can sample a functionally different mode, represents a more efficient use of the available budget than collecting a large number of samples from a single chain.
In Section~\ref{sec:efficient_sampling}, we propose an upper bound for MCMC chains necessary to observe all functionally diverse posterior modes, which is a key criterion for successful inference.
The perspective of parameter posterior symmetries thus lends a new theoretical justification to previous efforts in multi-chain MCMC that are motivated mainly by the exploitation of parallel computing resources \cite{rosenthal_2000_ParallelComputingMonte, margossian_2022_NestedHatAssessing}.
Our experiments in Section~\ref{sec:experiments} suggest that this approach is indeed more effective than single-chain MCMC in BNNs with many symmetries in their parameter posterior density.
In agreement with \citet{izmailov_what_2021}, our findings advocate to focus on function-space instead of parameter-space mixing during MCMC sampling.

We thus view the existence of equioutput parameter states as a benign phenomenon.
That said, there are still benefits to be gained from removing the symmetries: with a parameter posterior density reduced to the minimal parameter set sufficient to represent its full functional diversity, we get an opportunity for better interpretation, and possibly even analytical approximation.
We demonstrate the potential of symmetry removal in Section~\ref{subsec:results_symmetry_removal} by means of a custom algorithm (Supplementary Material~\ref{app:symmetry_removal}).
In the following section, we provide the mathematical background and introduce the characterization and formal notation of equioutput transformations.

\section{Background and Notation} \label{sec:background}
\subsubsection{MLP Architectures.}
In this work, we consider NNs of the following form. Let
$f: \mathcal{X} \to \mathcal{Y}$ represent an MLP with $K$ layers, where layer $l \in \{1,\ldots,K\}$ consists of $M_l$ neurons, mapping a feature vector $\bm{x} = (x_1,\ldots,x_n)^\top \in \mathcal{X} \subseteq \mathbb{R}^n, n \in \mathbb{N},$ to an outcome vector $$f(\bm{x}) =:\hat{\bm{y}} = (\hat{y}_1,\ldots,\hat{y}_m)^\top \in \mathcal{Y} \subseteq \mathbb{R}^m, ~~ m \in \mathbb{N},$$ to estimate $\bm{y} = (y_1,\ldots,y_m)^\top \in \mathcal{Y}$.
The $i$-th neuron in the $l$-th layer of the MLP is associated with the weights $w_{lij},~j=1,\ldots, M_{l-1},$ and the bias $b_{li}$.
We summarize all the MLP parameters in the vector $$\bm{\theta} :=  (w_{211},\ldots,w_{KM_KM_{K-1}},b_{21},\ldots,b_{KM_K})^\top \in \Theta \subseteq \mathbb{R}^d$$
and write $f_{\bm{\theta}}$ to make clear that the MLP is parameterized by $\bm{\theta}$.
For each hidden layer $l\in\{2,\ldots,K-1\}$, the inputs are linearly transformed and then activated by a function $a$.
More specifically, we define the
pre-activations 
of the $i$-th neuron in the $l$-th hidden layer as
$$o_{li} = \sum_{j = 1}^{M_{l-1}} w_{lij}z_{(l-1)j} + b_{li}$$ 
with post-activations
$z_{(l-1)i} = a(o_{(l-1)i})$ from the preceding layer.
For the input layer, we have $z_{1i} = x_i, i=1,\ldots,n$, and for the output layer,
$z_{Ki} = \hat{y}_i, i=1,\ldots,M_{K}$.

\subsubsection{Predictive Uncertainty.}

In the Bayesian paradigm, a prior density $p(\bm{\theta})$ is imposed on the parameters, typically as part of a Bayesian model of the data.
Using Bayes' rule, the parameter posterior density $$p(\bm{\theta} | \mathcal{D}) = \frac{p(\mathcal{D} | \bm{\theta}) p(\bm{\theta})}{p(\mathcal{D})} $$ updates this prior belief based on the information provided by the data $\mathcal{D}$ and encoded in the likelihood $p(\mathcal{D}|\bm{\theta})$.
In supervised learning, the data are typically given by a set of $N$ feature vectors $\bm{x} \in \mathcal{X}$ and outcome vectors $\bm{y} \in \mathcal{Y}$, forming the dataset $\mathcal{D} = \{(\bm{x}^{(1)}, \bm{y}^{(1)}),\ldots,(\bm{x}^{(N)}, \bm{y}^{(N)})\}$.
The PPD $p(\bm{y}^\ast | \bm{x}^\ast, \mathcal{D})$ quantifies the predictive or functional uncertainty of the model for a new observation $(\bm{x}^\ast, \bm{y}^\ast) \in \mathcal{X}\times\mathcal{Y}$.
Since $$p(\bm{y}^\ast | \bm{x}^\ast, \mathcal{D}) = \int_\Theta p(\bm{y}^\ast | \bm{x}^\ast, \bm{\theta}) p(\bm{\theta} | \mathcal{D}) \diff\bm{\theta},$$ deriving this uncertainty requires access to the posterior density $p(\bm{\theta} | \mathcal{D})$, which can be estimated from MCMC sampling.

\subsubsection{Equioutput Transformations.}
Let us now characterize the notion of equioutput parameter states, and the transformations to convert between them, more formally.
Two parameter states $\bm{\theta}, \bm{\theta}^\prime$ are considered \textit{equioutput} if the maps $f_{\bm{\theta}}, f_{\bm{\theta}^\prime}$ yield the same outputs for all possible inputs from $\mathcal{X}$.
We denote this equivalence relation (see proof in Supplementary Material~\ref{app:equiv_rel}) by $\sim$ and write:
$$\bm{\theta} \sim \bm{\theta}^\prime \Longleftrightarrow f_{\bm{\theta}}(\bm{x}) = f_{\bm{\theta}^\prime}(\bm{x}) \,\forall\,\bm{x}\in\mathcal{X}, ~~ \bm{\theta}, \bm{\theta}^\prime \in \Theta.$$
The equioutput relation is always defined with respect to a particular MLP $f$, which we omit in our notation when it is clear from the context.

All MLPs with more than one neuron in at least one hidden layer exhibit such equioutput parameter states that arise from permutation invariances of the input-output mapping \cite{hecht-nielsen_algebraic_1990, kurkova_functionally_1994}. 
Since the operations in the pre-activation of the $i$-th neuron in the $l$-th layer commute\footnote{
Recall that the pre-activation of neuron $i$ in layer $l$ is $o_{li} = \sum_{j = 1}^{M_{l-1}} w_{lij}z_{(l-1)j} + b_{li}$.
By the commutative property of sums, any permutation $\pi: J \rightarrow J$ of elements from the set $J = \{1, \dots, M_{l - 1} \}$ will lead to the same pre-activation:\\ $o_{li} = \sum_{j \in J} w_{lij}z_{(l-1)j} + b_{li} = \sum_{j \in \pi(J)} w_{lij}z_{(l-1)j} + b_{li}.$
}, the $M_l>1$ neurons of a hidden layer $l$ can be freely interchanged by permuting their associated parameters.
In addition, equioutput transformations can arise from the use of certain activation functions with inherent symmetry properties.
For example, in the case of tanh, the signs of corresponding parameters can be flipped using $\tanh(x) = -\tanh(-x)$.
For ReLU activations, a scaling transformation can be applied such that the mapping of the network remains unchanged, i.e., $\text{ReLU}(x) = c^{-1} \cdot \text{ReLU}(c \cdot x)$ for $|c| > 0$ (see also Supplementary Material~\ref{app:functional_equality}). 

We consider transformation maps that are linear in $\bm{\theta}$ and induce a finite amount of equioutput transformation matrices, which includes, for example, the tanh activation function.
The ReLU activation function allows for infinitely many possibilities of re-scaling the weights and is excluded from our findings.
More specifically, let $$\mathcal{F}_{\bm{T}}: \Theta \to \Theta, \bm{\theta} \mapsto \bm{T}\bm{\theta}, ~~ \bm{T} \in \mathbb{R}^{d \times d},$$ be an activation-related transformation of a parameter vector that might, for instance, encode an output-preserving sign flip. 
$\mathcal{F}_{\bm{T}}$ constitutes an \textit{equioutput} transformation if $f_{\bm{\theta}}(\cdot) = f_{\mathcal{F}_{\bm{T}}(\bm{\theta})}(\cdot)$.
We collect all output-preserving transformation matrices $\bm{T}$ in the set $\mathcal{T}$, i.e.,
$$\mathcal{T} = \left \{ \bm{T} \in \mathbb{R}^{d \times d} \; | \; f_{\bm{\theta}}(\cdot) = f_{\mathcal{F}_{\bm{T}}(\bm{\theta})}(\cdot) \right\}.$$
Similarly, let $$\mathcal{F}_{\bm{P}}: \Theta \to \Theta, \bm{\theta} \mapsto \bm{P}\bm{\theta}, ~~ \bm{P} \in \{0, 1\}^{d \times d},$$ be a transformation that permutes elements in the parameter vector. 
We define the set of permutation matrices that yield equioutput parameter states as
$$\mathcal{P} = \left \{ \bm{P} \in \mathbb{R}^{d \times d} \; | \; f_{\bm{\theta}}(\cdot) = f_{\mathcal{F}_{\bm{P}}(\bm{\theta})}(\cdot) \right\}.$$
The cardinality of $\mathcal{P}$ is at least $\prod_{l = 2}^{K - 1} M_l!$ \cite{hecht-nielsen_algebraic_1990} when traversing through the NN from the first layer in a sequential manner, applying to each layer permutations that compensate for permutations in its predecessor.

Since activation functions operate neuron-wise, activation- and permutation-related equioutput transformations do not interact (for instance, we could permute the associated weights of two neurons and later flip their sign).
We can, therefore, define arbitrary combinations of activation and permutation transformations as
$$\mathcal{E} = \left \{\bm{E} = \bm{T} \bm{P} \in \mathbb{R}^{d \times d}, \bm{T} \in \mathcal{T}, \bm{P} \in \mathcal{P} \; | \; f_{\bm{\theta}}(\cdot) = f_{\mathcal{F}_{\bm{E}}(\bm{\theta})}(\cdot) \right\}.$$
The transformation matrices in $\mathcal{E}$ will exhibit a block-diagonal structure with blocks corresponding to network layers. This is due to the permutations $\bm{P}$ affecting both incoming and outgoing weights, but only in the sense that two incoming and two outgoing weights swap places, never changing layers. The activation-related sign flips or re-scalings occur neuron-wise, making $\bm{T}$ a diagonal matrix that does not alter the block-diagonal structure of $\bm{P}$. 

\begin{figure}[ht]
	\centering
	\includegraphics[width=1.0\textwidth]{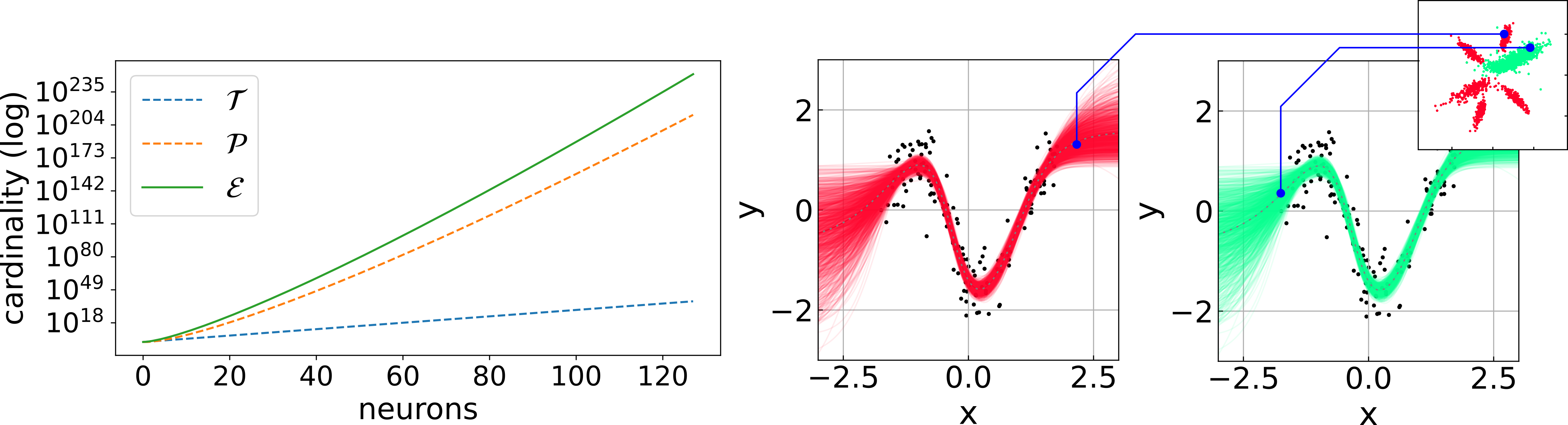}
	\caption{Example of tanh-activated MLPs. \textit{Left}: Cardinality lower bound of the equioutput transformation set for a single hidden layer with 1 to 128 neurons; the functional redundancy factor for a network with 128 neurons is $1.31 \cdot 10^{254}$. \textit{Right}: A ten-dimensional MLP parameter posterior (top-right corner, depicted as bivariate marginal density) exhibits symmetries, such that all red sample clusters are equioutput-related to the green cluster. The associated function spaces are identical, i.e., many posterior modes are redundant.}
	\label{fig:redundancy}
\end{figure}

For the cardinality of the set $\mathcal{E}$ of equioutput transformations, we can establish a lower bound that builds upon the minimum cardinality of $\mathcal{P}$: 
$$|\mathcal{E}| \geq \textstyle\prod_{l = 2}^{K - 1} M_l! \cdot |\mathcal{T}_l|,$$
where $|\mathcal{T}_l|$ denotes the number of activation-related transformations applicable to neurons in layer $l$.
From this, it becomes immediately clear that the amount of functional redundancy increases rapidly with the network size (see also Figure~\ref{fig:redundancy}).
As a result of equioutput parameter states, the MLP parameter posterior density exhibits functional redundancy in the form of symmetries for commonly used priors (see Section~\ref{sec:efficient_sampling}):
\begin{align}
	p(\bm{\theta} | \mathcal{D}) &= \frac{p(\mathcal{D}|\bm{\theta}) p(\bm{\theta})}{p(\mathcal{D})} \overset{\ref{formula:prior}, \ref{formula:likelihood}}{=} \frac{p(\mathcal{D}|\bm{\bm{E}\theta}) p(\bm{E}
		\bm{\theta})}{p(\mathcal{D})} = p(\bm{E}\bm{\theta} | \mathcal{D}), ~~
  \bm{\theta} \in \Theta, \bm{E} \in \mathcal{E}.
  \label{formula:posterior_equality}
\end{align}

\section{Efficient Sampling} \label{sec:efficient_sampling}
These symmetric structures in the parameter posterior density suggest that sampling can be made more efficient.
In the following section, we show that the PPD can theoretically be obtained from a small reference set of non-equivalent parameter states and propose an upper bound on Markov chains that suffice to sample all non-symmetric posterior modes.

\subsection{Posterior Reference Set} \label{subsec:refset}
As introduced in Section \ref{sec:background}, for each parameter state $\bm{\theta}$ of an NN, there are functionally redundant counterparts $\bm{\theta}^\prime$ related to $\bm{\theta}$ by an equioutput transformation, such that $f_{\bm{\theta}}(\cdot) = f_{\bm{\theta}^\prime}(\cdot)$.
We can use this equivalence relation to dissect 
the parameter space $\Theta$ into disjoint equivalence classes.
For this, let the \textit{reference set $\mathcal{S}_1$} be a minimal set of representatives of each equivalence class (cf. \emph{open minimal sufficient search sets} in \cite{chen_geometry_1993}).
All parameter states in $\mathcal{S}_1$ are functionally diverse, i.e., $\bm{\theta}, \tilde{\bm{\theta}} \in \mathcal{S}_1 \Rightarrow \bm{\theta} \not\sim \tilde{\bm{\theta}}$, and each element in $\Theta$ is equivalent to exactly one element in $\mathcal{S}_1$.
For a finite amount of equioutput transformations, as in the case of tanh-activated MLPs (finite possibilities of sign-flip combinations of hidden neurons), the NN parameter space can then be dissected into $|\mathcal{E}|$ disjoint \textit{representative sets}, which contain equioutput transformations of the elements of the reference set, in the following way.

\begin{proposition}[Parameter space dissection] \label{prop:param_space_dissection}
Let $\mathcal{S}_1$ be the reference set of uniquely identified network parameter states. Then, for a finite number of equioutput transformations, it holds that the parameter space can be dissected into $|\mathcal{E}|$ disjoint, non-empty representative sets up to a set $\mathcal{S}^0 \subset \Theta$, i.e.,
\begin{equation}
\Theta = \left( \textstyle\ \dot \bigcup_{j=1}^{|\mathcal{E}|} \mathcal{S}_j \right) \dot \cup \, \mathcal{S}^0 \mbox{, where } \mathcal{S}_j \cong \left \{\bm{\theta} ~|~ \bm{\theta} = \bm{E}_j \bm{\theta}' \quad \forall \bm{\theta}' \in \mathcal{S}_1, \bm{E}_j \in \mathcal{E} \right\}, \label{formula:union}
\end{equation}
where $\dot \cup$ denotes the union over disjoint sets.
We use $\mathcal{S}^0$ as a residual quantity to account for cases that cannot be assigned unambiguously to one of the sets $\mathcal{S}_j$ because they remain unchanged even under a transformation with non-identity matrices $\bm{E}_j \in \mathcal{E}$.
\end{proposition}

The edge cases that make up $\mathcal{S}^0$ exist, e.g., on the boundary of two classes \cite{chen_geometry_1993} or in degenerated cases such as the zero vector \cite{sussmann_uniqueness_1992, VLACIC2021107485}.
For a characterization of the involved sets, as well as a proof sketch, see Supplementary Material\ref{proof_dissection}.

Equioutput parameter states have the same posterior probabilities $p(\bm{\theta} | \mathcal{D}) = p(\bm{E}\bm{\theta} | \mathcal{D})$ if the prior is transformation-invariant; see Supplementary Material~\ref{app:posteriors}, Equations~(\ref{formula:prior})-(\ref{formula:likelihood}). Moreover, equioutput parameter states produce by definition the same predictions $p(\bm{y}^\ast | \bm{x}^\ast, \bm{\theta}) = p(\bm{y}^\ast | \bm{x}^\ast, \bm{E} \bm{\theta})$ for any $\bm{E} \in \mathcal{E}$. Thus, the following corollary holds.

\begin{corollary}[Reformulated posterior predictive density] \label{cor:predictive}
Let $\mathcal{E}$ be finite. As in Proposition~\ref{prop:param_space_dissection}, consider the disjoint non-empty sets $\mathcal{S}_j, j \in \{1, \dots, |\mathcal{E}|\}$, and residual space $\mathcal{S}^0$.
If the prior density $p(\bm{\theta})$ is trans\-for\-mation-invariant,
then the posterior predictive density expresses as
\begin{align}
    p(\bm{y}^\ast | \bm{x}^\ast, \mathcal{D}) &= \int_{\Theta} p(\bm{y}^\ast | \bm{x}^\ast, \bm{\theta}) p(\bm{\theta} | \mathcal{D}) \diff\bm{\theta} \label{eq:ppd}
    \\
    &= |\mathcal{E}| \int_{\mathcal{S}_j} p(\bm{y}^\ast | \bm{x}^\ast, \bm{\theta})p(\bm{\theta}| \mathcal{D}) \diff \bm{\theta} 
    + \int_{\mathcal{S}^0} p(\bm{y}^\ast | \bm{x}^\ast, \bm{\theta})p(\bm{\theta}| \mathcal{D}) \diff \bm{\theta} \notag \\
    &\approx |\mathcal{E}| \int_{\mathcal{S}_j} p(\bm{y}^\ast | \bm{x}^\ast, \bm{\theta})p(\bm{\theta}| \mathcal{D}) \diff \bm{\theta}. \label{eq:reduced_predictive} 
\end{align}
\end{corollary}

The proof of Corollary~\ref{cor:predictive} is given in Supplementary Material~\ref{app:posteriors}.
It follows from Proposition~\ref{prop:param_space_dissection} and the assumption of transformation-invariant prior densities, which is often satisfied in practice (e.g., for widely-applied isotropic Gaussian priors). We can further approximate (\ref{eq:ppd}) by (\ref{eq:reduced_predictive}) as the set $\mathcal{S}^0\subset\mathbb{R}^d$ is of negligible size (depending on $\Theta$, potentially even with zero Lebesgue measure). 

As a consequence of Corollary~\ref{cor:predictive}, the PPD can be obtained up to the residual set by only integrating over uniquely identified parameter states from one of the sets $\mathcal{S}_j$, with a multiplicative factor $|\mathcal{E}|$ that corrects the probability values by the amount of redundancy in the posterior.
In other words, only a fraction $1 / |\mathcal{E}|$ of the posterior must be sampled in order to infer a set of uniquely identified parameter states of the NN, and thus, to obtain the full PPD.
This reduces the target sampling space drastically, as illustrated in Figure~\ref{fig:redundancy}.
For example, it allows the posterior space of a single-layer, tanh-activated network with 128 neurons to be effectively reduced to a $10^{254}$-th of its original size.

In the case of an infinite amount of equioutput transformations, such as in ReLU-activated MLPs (the scaling factor $|c| > 0$ can be chosen arbitrarily), we can use similar reasoning.
Only one representative set of the posterior density needs to be observed in order to capture the full functional diversity of a network because the integrals over two representative sets are identical.
For a more in-depth discussion of ReLU symmetries, see, for example, \cite{bona-pellissier_2021_ParameterIdentifiabilityDeep}.

\subsubsection{How to Obtain a Representative Set?}
In practice, when using Monte Carlo to approximate Equation~(\ref{eq:reduced_predictive}), it is not necessary to actually constrain the sampling procedure to a specific set $\mathcal{S}_j$, which might indeed not be straightforward\footnote{
\citet{ensign_2017_ComplexityExplainingNeural} demonstrate that finding invariant representations for groups acting on the input space is an NP-hard problem. 
While we are not aware of such a result for the parameter space, the NP-hardness in \cite{ensign_2017_ComplexityExplainingNeural} for permutations of the inputs only suggests a similar property in our case.
}.
Since any equioutput transformation is known \textit{a priori}, we just need to be aware of the fact that each sample can theoretically be mapped to different representative sets after running the sampling procedure.
Hence, for the calculation of the PPD integral, the samples can remain scattered across the various representative sets as long as they cover all functionally diverse parameter states.
For the purpose of providing better interpretability and analytic approximation of the posterior, it may still be worthwhile to explicitly remove the symmetries.
In Section~\ref{subsec:results_symmetry_removal}, we demonstrate such symmetry removal using a custom algorithm for tanh-activated networks (Supplementary Material~\ref{app:symmetry_removal}).

\subsection{An Upper Bound for Markov Chains} \label{subsec:upperbound}
The question remains how many samples are needed to approximate a set of uniquely identified parameter states sufficiently well.
Even in a symmetry-free setting, BNN posteriors can exhibit multiple functionally diverse modes representing structurally different hypotheses, depending on the network architecture and the underlying data-generating process.
For example, in Section~\ref{subsec:results_symmetry_removal}, we discuss the case of an under-parameterized network that preserves three distinctive modes caused by its restricted capacity.

In the following, we assume $\nu \in \mathbb{N}$ functionally diverse modes with the goal of visiting every mode or its local proximity at least once when running MCMC.
As the ability to switch from one mode to another within a chain depends on various factors, such as the acceptance probability and the current state of other parameters, increasing the number of samples per chain does not necessarily correlate with the number of visited modes.
We, therefore, propose to focus on the number of independent chains, rather than the number of samples per chain, to effectively control the number of visited modes.

This further allows us to derive an upper bound for the number of independent chains that are required to visit every mode at least once.
The number of samples from each chain will then ultimately determine the approximation quality.
In the computation of the PPD, we formulate the Monte Carlo integration over all samples from all chains simultaneously \cite{margossian_2022_NestedHatAssessing}.
In practice, given a user-defined number of maximal resources $\rho$ (e.g., CPU cores), the following proposition provides a lower bound on the probability that the number of chains $\mathcal{G}$ necessary to visit every mode remains below the resource limit of the user (i.e., $\mathcal{G} < \rho$).

\begin{proposition}[Probabilistic bound for sufficient number of Markov chains] \label{prop:upper_bound}
Let $\pi_1, \ldots,\pi_\nu$ be the respective probabilities of the $\nu$ functionally diverse modes to be visited by an independently started Markov chain and $\Pi_J := \sum_{j \in J} \pi_j$. Then, given $\rho$ chains,
\begin{equation}
  \mathbb{P}(\mathcal{G} < \rho) \geq 1 - \rho^{-1} \left\{\textstyle \sum_{q = 0}^{\nu - 1}(-1)^{\nu - 1 - q} \sum_{J: |J| = q} (1 - \Pi_J)^{-1}\right\} \label{formula:ccp}.  
\end{equation}
\end{proposition}

The proof can be found in Supplementary Material~\ref{app:upper}.
Note that this bound is independent of the NN architecture and only depends on the assumptions about the number and probabilities of functionally diverse modes $\nu$, disregarding symmetric copies.
Proposition~\ref{prop:upper_bound} can be used to calculate the number of MCMC chains given certain assumptions -- for example, from domain knowledge, or in a worst-case scenario calculation -- and thus provides practical guidance for MCMC sampling of MLPs.
Judging by the comparably high predictive performance of local approximations such as LA and DE \cite{lakshminarayanan_simple_2017, mackay_1992_BayesianInterpolation}, we conclude that a small amount of functional modes is reasonable to assume in practice. 
Our qualitative experiments in Section~\ref{sec:experiments} support this supposition.

As an example of applying Proposition~\ref{prop:upper_bound}, assume $\nu = 3$ functionally diverse modes in a reference set with $\pi_1 = 0.57, \pi_2 = 0.35, \pi_3 = 0.08$ (chosen to represent a rather diverse functional mode set).
An upper bound of $\rho = 1274$ chains ensures that we observe all functionally diverse modes
with probability $\mathbb{P}(\mathcal{G} < \rho) \geq 0.99$.

\section{Experiments} \label{sec:experiments}
We now investigate our theoretical findings and compare the resulting approach to single-chain MCMC and DE.
In all experiments\footnote{
\scriptsize
\href{https://anonymous.4open.science/r/efficient_mcmc_sampling_in_bnns_symmetry-6FAF/}{https://anonymous.4open.science/r/efficient\_mcmc\_sampling\_in\_bnns\_symmetry-6FAF/}
}, we employ a Bayesian regression model with a normal likelihood function, standard normal prior for parameters $\bm{\theta}$, and a truncated standard normal prior restricted to the positive real line for the variance of the normal likelihood, which we treat as a nuisance parameter.
Depending on the task, we either use a No-U-Turn sampler \cite{hoffman_no-u-turn_2014} with $2^{10}$ warmup steps to collect a single sample from the posterior or derive the maximum-a-posteriori estimator using a gradient-based method (details are given in Supplementary Material~\ref{app:mcmc},~\ref{app:point_estimates}).

\subsection{Performance Comparison} \label{subsec:results_performance}
\begin{table}[t]
    \caption{Mean log pointwise predictive density (LPPD) values on test sets (larger is better; one standard error in parentheses). The highest performance per dataset and network is highlighted in bold.}
    \resizebox{1.0\textwidth}{!}{
    \begin{tabular}{l|rrr|rrr}
    \multicolumn{1}{c}{} & \multicolumn{3}{c}{\textbf{Smaller network $f_1$}} & \multicolumn{3}{c}{\textbf{Larger network $f_2$}} \\
    \multicolumn{1}{c}{} & \multicolumn{1}{c}{MCMC (ours)} & \multicolumn{1}{c}{MCMC (s.c.)} & \multicolumn{1}{c}{DE} & \multicolumn{1}{c}{MCMC (ours)} & \multicolumn{1}{c}{MCMC (s.c.)} & \multicolumn{1}{c}{DE} \\ \hline
    $\mathcal{D}_S$ & \textbf{-0.53} ($\pm$ 0.09) & -0.56 ($\pm$ 0.11) & -0.58 ($\pm$ 0.11) & \textbf{-0.59} ($\pm$ 0.12) & \textbf{-0.59} ($\pm$ 0.12) & -2.13 ($\pm$ 0.03) \\
    $\mathcal{D}_I$ & \textbf{0.79} ($\pm$ 0.06) & 0.65 ($\pm$ 0.07) & 0.56 ($\pm$ 0.06) & \textbf{0.91} ($\pm$ 0.09) & \textbf{0.91} ($\pm$ 0.09) & -2.02 ($\pm$ 0.02) \\  
    $\mathcal{D}_R$ & 0.64 ($\pm$ 0.10) & \textbf{0.75} ($\pm$ 0.11) & -1.46 ($\pm$ 0.06) & \textbf{0.95} ($\pm$ 0.08) & \textbf{0.95} ($\pm$ 0.08) & -2.20 ($\pm$ 0.02) \\ 
    Airfoil & \textbf{-0.74} ($\pm$ 0.04) & -0.80 ($\pm$ 0.05) & -1.62 ($\pm$ 0.03) & \textbf{0.92}  ($\pm$ 0.05) & 0.72 ($\pm$ 0.10) & -2.17 ($\pm$ 0.01) \\ 
    Concrete & \textbf{-0.41} ($\pm$ 0.05) & -0.44 ($\pm$ 0.06) & -1.59 ($\pm$ 0.03) & \textbf{0.26} ($\pm$ 0.07) & 0.25 ($\pm$ 0.07) & -2.03 ($\pm$ 0.01) \\ 
    Diabetes & \textbf{-1.20} ($\pm$ 0.07) & \textbf{-1.20} ($\pm$ 0.07) & -1.47 ($\pm$ 0.07) & \textbf{-1.18} ($\pm$ 0.08) & -1.22 ($\pm$ 0.09) & -2.09 ($\pm$ 0.04) \\ 
    Energy & \textbf{0.92} ($\pm$ 0.04) & 0.69 ($\pm$ 0.12) & -1.76 ($\pm$ 0.02) & 2.07 ($\pm$ 0.46) & \textbf{2.38} ($\pm$ 0.11) & -1.99 ($\pm$ 0.02) \\
    ForestF & \textbf{-1.37} ($\pm$ 0.07) & \textbf{-1.37} ($\pm$ 0.07) & -1.60 ($\pm$ 0.06) & \textbf{-1.43} ($\pm$ 0.45) & -1.69 ($\pm$ 0.49) & -2.20 ($\pm$ 0.02) \\ 
    Yacht & \textbf{1.90} ($\pm$ 0.16) & 1.29 ($\pm$ 0.56) & -1.14 ($\pm$ 0.14) & \textbf{3.31} ($\pm$ 0.21) & 0.15 ($\pm$ 0.09) & -2.18 ($\pm$ 0.03) 
    \end{tabular}
    }
    \label{tab:result_posteriors}
\end{table}

In our first experiment, we demonstrate the predictive performance of BNNs, where the PPD is calculated based on MCMC sampling, using the derived upper bound for the number of chains (ours). In this case, we collect one sample per chain for $G$ chains, and thus $G$ samples in total.
This is compared to MCMC sampling collecting $G$ samples from a single chain (s.c.), and DE with ten ensemble members on three synthetic datasets ($\mathcal{D}_S$, $\mathcal{D}_I$, and $\mathcal{D}_R$) as well as benchmark data from \citet{Dua.2019} (for dataset details and additional results on LA, see Supplementary Material~\ref{app:datasets} and~\ref{app:results_performance}, respectively).
We use a smaller NN $f_1$ with a single hidden layer containing three neurons and a larger network $f_2$ with three hidden layers having 16 neurons each, both with tanh activation.
As in Section~\ref{subsec:upperbound}, we assume three functionally diverse modes $\nu=3$ and mode probabilities $\pi_1 = 0.57, \pi_2 = 0.35, \pi_3 = 0.08$ as in the given example.
To demonstrate the performance of our MCMC-based PPD approximation,
we measure the goodness-of-fit on the test data
using the log point-wise predictive density (LPPD; \cite{gelman_understanding_2013})
\begin{equation}
\label{eq:lppd}
\text{LPPD} =
\log \int_\Theta p(\bm{y}^\ast | \bm{x}^\ast, \bm{\theta}) p(\bm{\theta} | \mathcal{D})
\diff\bm{\theta}
\approx
\log {
\left(
\displaystyle\tfrac{1}{G} \sum_{g=1}^G p(\bm{y}^\ast | \bm{x}^\ast, \bm{\theta}^{(g)})
\right)},
\end{equation}
where $\bm{\theta}^{(1)},\ldots,\bm{\theta}^{(G)}$ are $G$ samples obtained across all chains via MCMC sampling
from the parameter posterior density $p(\bm{\theta} | \mathcal{D})$.
Equation~(\ref{eq:lppd}) is evaluated at each test point $({\bm{x}^\ast}, {\bm{y}^\ast})$.
Table~\ref{tab:result_posteriors} reports the mean LPPD across
$N^{\ast}$ independent test points for each combination
of dataset and sampling scheme (see Appendix~\ref{app:datasets} for details).
Our results clearly indicate that using only a moderate amount of Markov chains, following our approach, yields equal or even better performance than single-chain MCMC and DE in all but two experiments.

\subsection{Practical Evaluation of Corollary~\ref{cor:predictive}} \label{subsec:results_corollary}
\begin{figure}[ht]
	\includegraphics[width=1.0\textwidth]{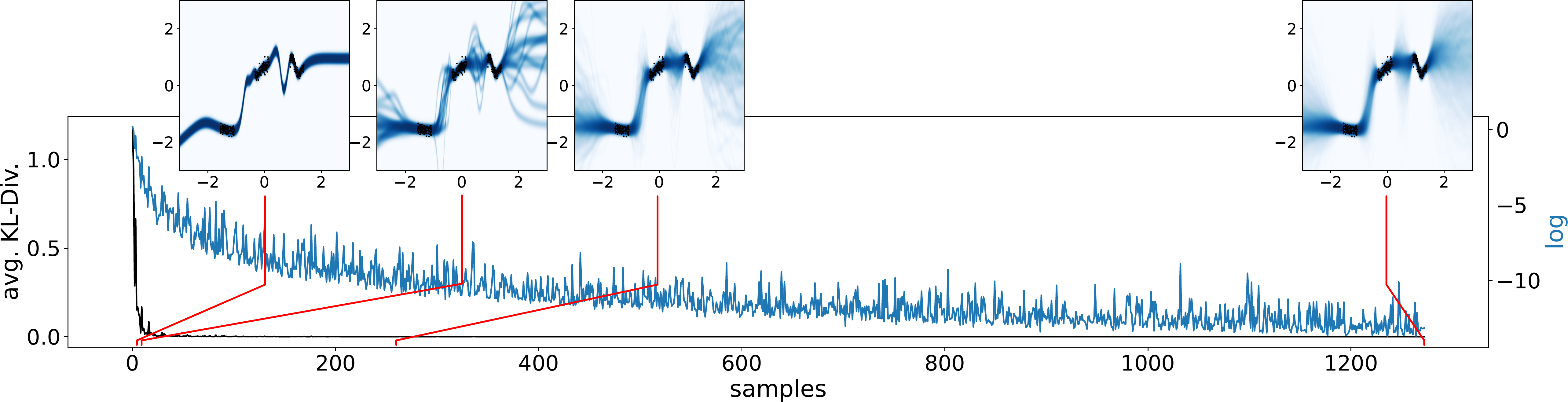}
	\caption{Convergence of MCMC depicted as the change in KL-divergence on original (black) and log-scale (blue) when consecutively adding another sample from a new and independent chain and re-estimating the posterior density. Small overlaying plots: approximated PPD of the network after $2^0$, $2^4$, $2^8$, and $G = 1274$ samples; darker colors correspond to higher probabilities.}
	\label{fig:result_posteriors}
\end{figure}

Next, we investigate the property derived in Corollary~\ref{cor:predictive} using our proposed upper bound of chains, again with the assumption from the example in Section~\ref{subsec:upperbound}.
To this end, we analyze the PPD for dataset $\mathcal{D}_I$, using network $f_2$.
For every newly collected sample in the MCMC run, the updated PPD is computed approximately on a two-dimensional (input/output) grid.
Then, the Kullback-Leibler (KL) divergence between consecutive densities is averaged over the grid of input values of $f_2$ (details in Supplementary Material~\ref{app:kl_divergence}).
As shown in Figure~\ref{fig:result_posteriors}, despite the size of the network $f_2$ and the high amount of equioutput parameter states $|\mathcal{E}| = \left(16! \cdot 2^{16}\right)^3 \approx 2.58 \cdot 10^{54}$, the PPD converges after notably fewer than $|\mathcal{E}|$ samples and plots of the function space indicate the saturation of functional diversity already after 1274 samples from as many chains. 

\subsection{Posterior Symmetry Removal} \label{subsec:results_symmetry_removal}
So far we were mainly concerned with the predictive performance of MCMC sampling.
Yet, mapping all samples to a joint representative set, as characterized in Section~\ref{subsec:refset}, has the potential to reduce the effective weight space enormously, facilitating interpretability and possibly even analytical approximation.
For this, we propose a custom algorithm for tanh-activated MLPs as a proof-of-concept.
Our algorithm removes symmetries in a data-dependent manner and thus minimizes the number of remaining modes in the representative set (details in Supplementary Material~\ref{app:symmetry_removal}).

\begin{figure}[H]
	\includegraphics[width=0.6\textwidth]{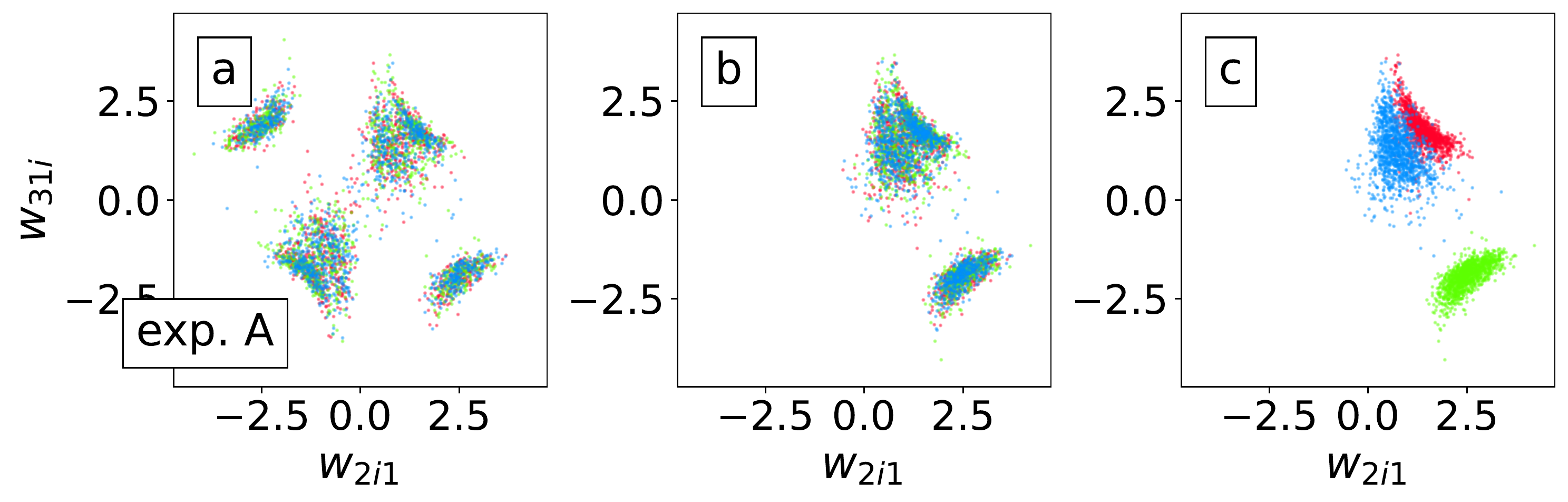}
	\includegraphics[width=0.4\textwidth]{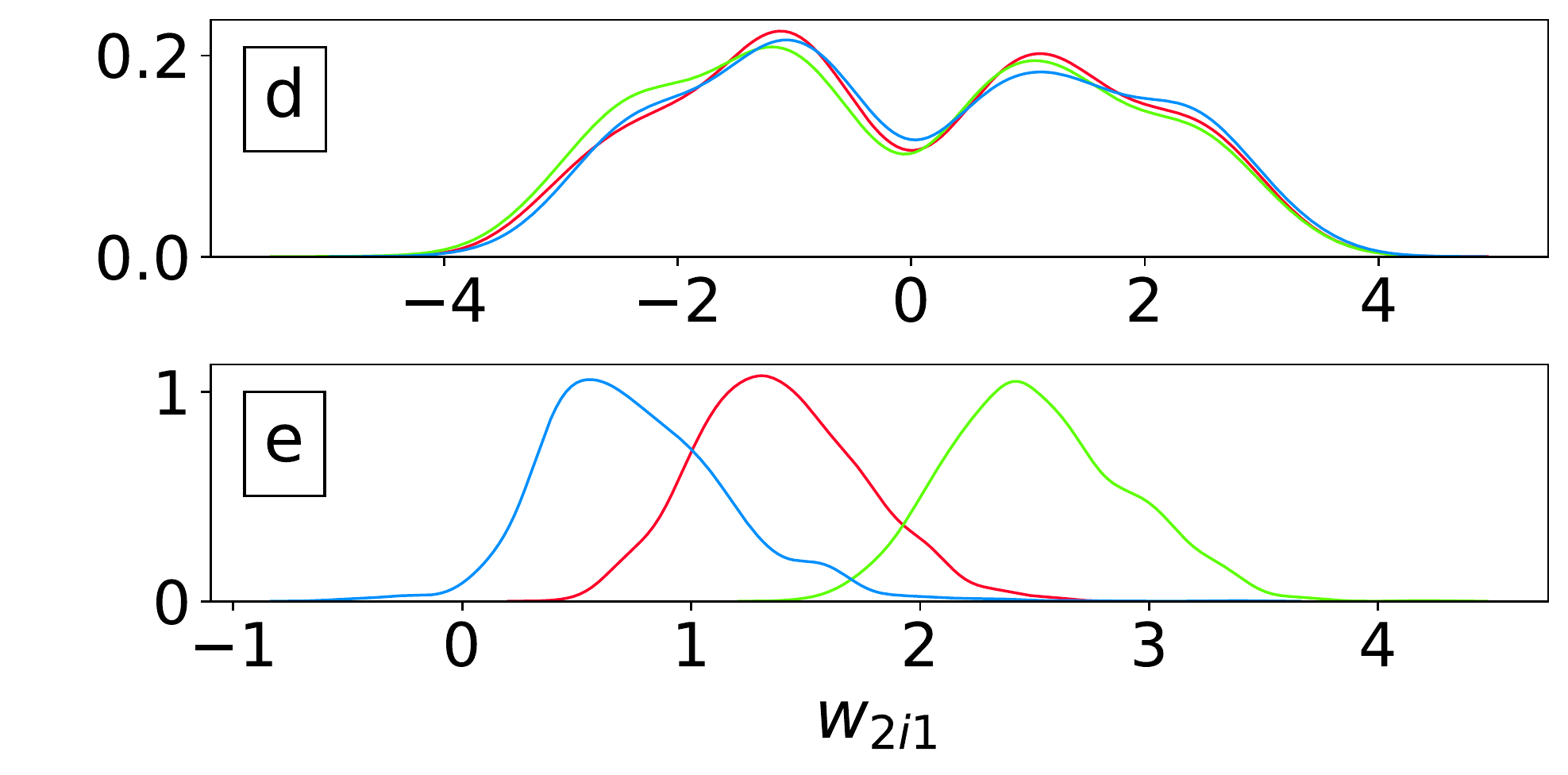}
	\includegraphics[width=1.0\textwidth]{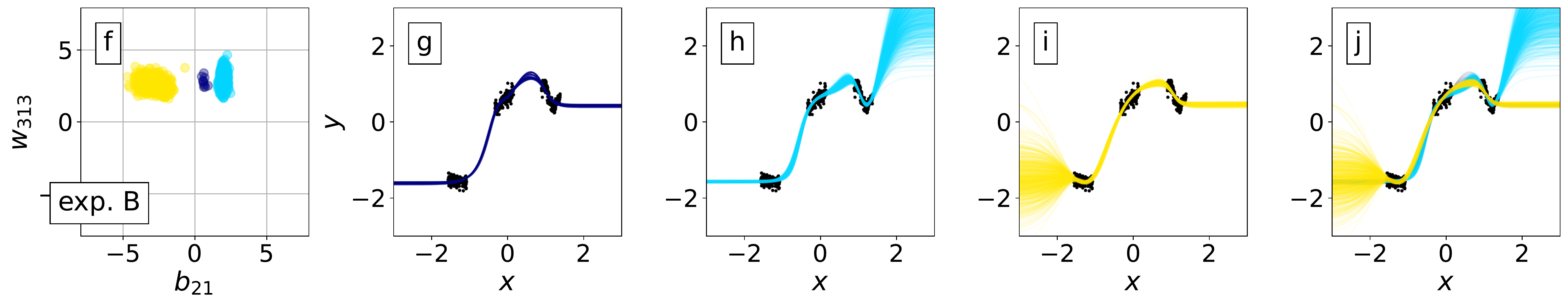}
	\caption{For experiment A (top row), the parameter subspaces of hidden neurons are visualized in their initial state as obtained when running MCMC (a), and after their transformation and reassignment during the steps of Algorithm~\ref{algo:complete_removal} (b, c). Different colors encode the current neuron index in the hidden layer. By reassigning neurons, Algorithm~\ref{algo:complete_removal} effectively finds an optimal reference set and allows to separate the multi-modal complex univariate marginal density (d) into three univariate densities.\\
    For experiment B (bottom row), the MLP parameter posterior density after symmetry removal results in a tri-modal system, here illustrated as a bivariate plot (f). Investigating these modes, we find that all are functionally diverse, i.e., represent a different hypothesis of the dataset (g-i). Combined, they form the full function space (j).}
	\label{fig:symmetry_removal}
\end{figure}

We demonstrate the efficacy of the approach for $f_1$ in two experiments A and B for datasets $\mathcal{D}_S$ and $\mathcal{D}_I$, respectively.
For experiment A (Figure~\ref{fig:symmetry_removal}, top) we visualize the neuron parameter space along the steps of the proposed algorithm (for details, see Algorithms~\ref{algo:tanh_removal}-\ref{algo:complete_removal} in Supplementary Material \ref{app:symmetry_removal}).
Different colors encode the current neuron index (i.e., one of $\{1, 2, 3\}$) in the hidden layer of the respective neuron parameter vector.
Initially, symmetries of the posterior densities are clearly noticeable (Figure~\ref{fig:symmetry_removal}a), and the neuron parameter vectors are distributed identically (Figure~\ref{fig:symmetry_removal}d).
Upon the first step of the algorithm (Figure~\ref{fig:symmetry_removal}b), the parameter space is effectively halved as a consequence of removing the tanh-induced sign-flip symmetry, and three clusters remain.
The second (clustering) step removes all permutation symmetries from the full posterior density by assigning areas of this parameter subspace to the neurons of the model.
This is depicted in Figure~\ref{fig:symmetry_removal}c and clearly shows the separation of states by different cluster colors.
In Figures~\ref{fig:symmetry_removal}d and ~\ref{fig:symmetry_removal}e, the univariate marginal density of each neuron's incoming weight $w_{2i1}$ reveals their reassignment to the parameter space.
After running our algorithm, each neuron exhibits a distinct unimodal density, resulting in a unimodal density in the full parameter space (visualization in Supplementary Material~\ref{app:results_full_posterior}).
We can conclude that only one functionally diverse mode exists in this case, which should also be recovered by local approximations like LA.

For experiment B (Figure~\ref{fig:symmetry_removal}, bottom), we focus on the visualization of the full parameter posterior density obtained after the application of the symmetry removal algorithm (Figure~\ref{fig:symmetry_removal}f). 
Three functionally diverse modes, represented by different colors, remain in the BNN posterior.
We can now interpret these modes in function space by further clustering the transformed samples using a spectral clustering approach (details in Supplementary Material~\ref{app:spectral_clustering}).
Figures~\ref{fig:symmetry_removal}g-j visualize the network parameter states in the function space, revealing three functionally diverse hypotheses the network can potentially learn from the given dataset.
Such knowledge allows for a better-suited approximation by, e.g., a mixture of Laplace approximations (MoLA; \cite{eschenhagen_2021_MixturesLaplaceApproximations}), as shown in Supplementary Material~\ref{app:results_application_example}.
Note that approaches focusing on a single mode, such as standard LA, would have captured only a third of the functional diversity. 
For meaningful UQ, it is thus imperative that all functionally diverse modes are accounted for.

\section{Discussion} \label{sec:discussion}
We showed that the PPD for Bayesian MLPs can be obtained from just a fraction of the parameter space due to the existence of equioutput parameter states.
Together with an upper bound on the number of MCMC chains to guarantee the recovery of every functionally diverse mode, our approach paves the way towards exact uncertainty quantification (up to a Monte Carlo error) in deep learning.
Furthermore, we demonstrate the use of symmetry removal and present a proof-of-concept approach to map samples of tanh-activated MLPs to a representative set.
This \textit{post-hoc} procedure improves the interpretability of the symmetry-free posterior density drastically and facilitates analytical approximations.
%
As a future research direction, we plan to investigate whether our MCMC sampling approach can be improved by initializing the sampling states in an informative way via ensemble training, building upon insights in \cite{wilson2020, pearce2020, graf2007}. 

\subsubsection*{Acknowledgments.}
LW is supported by the DAAD programme Konrad Zuse Schools of Excellence in Artificial Intelligence, sponsored by the German Federal Ministry of Education and Research.

\bibliographystyle{splncs04}
\bibliography{main}
\clearpage

\appendix
\section{Derivations} \label{app:derivations}
\subsection{Functional Equality of Equioutput Parameter States} \label{app:functional_equality}
We consider permutation and activation function-related equioutput transformation of the parameter states.
The following shows the functional equality for the interchange of neurons and two activation functions, tanh and ReLU.

\subsubsection{Permutation-Related Equality.}
Due to the commutative property of addition, the permutation of two neurons $v, w$ in the $(l-1)$-th layer does not change the pre-activation $o_{li}$ of the $i$-th neuron in the $l$-th layer.
\begin{alignat*}{3}
 	&o_{li} &&= (w_{li1} z_{(l-1)1} + b_{l1}) &&+ \dots + (w_{liv} z_{(l-1)v} + b_{lv}) + (w_{liw} z_{(l-1)w} + b_{lw}) \\
 	& && &&+ \dots + (w_{liM_{l-1}} z_{(l-1)M_{l-1}} + b_{lM_{l-1}}) \\
 	& &&= (w_{li1} z_{(l-1)1} + b_{l1}) &&+ \dots + (w_{liw} z_{(l-1)w} + b_{lw}) + (w_{liv} z_{(l-1)v} + b_{lv}) \\
 	& && &&+ \dots + (w_{liM_{l-1}} z_{(l-1)M_{l-1}} + b_{lM_{l-1}}) 
 \end{alignat*}

\subsubsection{Tanh-Related Equality.}
Since tanh is an odd function, flipping the signs of all parameters tethered to the outputs of incoming neurons (from layer $l-1$) and outgoing neurons (to layer $l+1$) as well as the associated bias parameter of layer $l$ leaves the output unchanged.
 \begin{align*}
 	w_{(l+1)ki} \tanh(o_{li}) &= w_{(l+1)ki} \tanh\left(\sum_{j = 1}^{M_{l-1}} w_{lij}z_{(l-1)j} + b_{li}\right) \\ 
    &= -w_{(l+1)ki} \tanh\left(\sum_{j = 1}^{M_{l-1}} -w_{lij}z_{(l-1)j} - b_{li}\right)\\
     &= -w_{(l+1)ki} \tanh(-o_{li}) 
\end{align*}

\subsubsection{ReLU-Related Equality.}
Similar to tanh, in the case of ReLU, rescaling the parameters of incoming (from layer $l-1$) and outgoing neuron outputs (to layers $l+1$) and the bias parameter of layer $l$ generates equioutput parameter states.
\begin{align*}
    w_{(l+1)ki} \text{ReLU}(o_{li}) &= w_{(l+1)ki} \text{ReLU}\left(\sum_{j = 1}^{M_{l-1}} w_{lij}z_{(l-1)j} + b_{li}\right) \\ 
    &= c \cdot w_{(l+1)ki} \text{ReLU}\left(\sum_{j = 1}^{M_{l-1}} c^{-1} \cdot w_{lij}z_{(l-1)j} + c^{-1} \cdot b_{li}\right)\\
    &= c \cdot w_{(l+1)ki} \text{ReLU}(c^{-1} \cdot o_{li}) 
\end{align*}

\subsection{Equioutput Equivalence Relation} \label{app:equiv_rel}
Let $f: \mathcal{X} \to \mathcal{Y}$ represent an MLP mapping a feature vector $\bm{x} \in \mathcal{X} \subseteq \mathbb{R}^n, n \in \mathbb{N},$ to an outcome vector $f(\bm{x}) \in \mathcal{Y} \subseteq \mathbb{R}^m$, $m \in \mathbb{N}$.
Two parameter states $\bm{\theta}, \bm{\theta}^\prime \in \Theta$ are considered \textit{equioutput} if the maps $f_{\bm{\theta}}, f_{\bm{\theta}^\prime}$ yield the same outputs for all possible inputs.
We denote this relation by $\sim$ and write:
$$\bm{\theta} \sim \bm{\theta}^\prime \Leftrightarrow f_{\bm{\theta}}(\bm{x}) = f_{\bm{\theta}^\prime}(\bm{x}) \,\forall\,\bm{x}\in\mathcal{X}, ~~ \bm{\theta}, \bm{\theta}^\prime \in \Theta.$$
$\sim$ constitutes an equivalence relation, which follows immediately from equality being an equivalence relation:
\begin{itemize}
    \item[E1] (Reflexivity) For $\bm{\theta}\in \Theta$, $\bm{\theta} \sim \bm{\theta}$ since $f_{\bm{\theta}}(\bm{x}) = f_{\bm{\theta}}(\bm{x}) \,\forall\,\bm{x}\in\mathcal{X}$.
    \item[E2] (Symmetry) For $\bm{\theta}, \bm{\theta}^\prime \in \Theta$, $$\bm{\theta} \sim \bm{\theta}^\prime \Leftrightarrow f_{\bm{\theta}}(\bm{x}) = f_{\bm{\theta}^\prime}(\bm{x}) \,\forall\,\bm{x}\in\mathcal{X} \Rightarrow f_{\bm{\theta}^\prime}(\bm{x}) = f_{\bm{\theta}}(\bm{x}) \,\forall\,\bm{x}\in\mathcal{X} \Leftrightarrow \bm{\theta}^\prime \sim \bm{\theta}$$
    \item[E3] (Transitivity) Let $\bm{\theta}_1 \sim \bm{\theta}_2$ and $\bm{\theta}_2 \sim \bm{\theta}_3$ for $\bm{\theta}_1, \bm{\theta}_2, \bm{\theta}_3 \in \Theta$. 
    Since 
    $$\bm{\theta}_1 \sim \bm{\theta}_2 \Leftrightarrow f_{\bm{\theta}_1}(\bm{x}) = f_{\bm{\theta}_2}(\bm{x})\,\forall\,\bm{x}\in\mathcal{X}$$
    and
    $$\bm{\theta}_2 \sim \bm{\theta}_3 \Leftrightarrow f_{\bm{\theta}_2}(\bm{x}) = f_{\bm{\theta}_3}(\bm{x})\,\forall\,\bm{x}\in\mathcal{X},$$
    we have, by the transitivity of equality,
    $$f_{\bm{\theta}_1}(\bm{x}) = f_{\bm{\theta}_3}(\bm{x})\,\forall\,\bm{x}\in\mathcal{X} \Leftrightarrow \bm{\theta}_1 \sim \bm{\theta}_3.$$
\end{itemize}

\subsection{Posterior Density Reference Set} \label{app:posteriors}

\subsubsection{Prior Invariance} \label{prior_inv}

For a neural network parameter state $\bm{\theta} \in \Theta \subseteq \mathbb{R}^d$ and any equioutput parameter state $\bm{E}_j \in \mathcal{E}$, the prior probability for $\bm{\theta}$ and $\bm{E}_j\bm{\theta}$ is identical if we assume the prior to be invariant under transformations $\mathcal{F}_{\bm{E}_j}: \Theta \to \Theta, \bm{\theta} \mapsto \bm{E}_j\bm{\theta}, \bm{E}_j \in \mathcal{E}$, i.e., $p(\bm{\theta}) = p(\bm{E}_j\bm{\theta})$.
Since equioutput parameter states produce, by definition, the same functional mapping, it is only reasonable to assign them identical prior probability.
A notable example is the universally-used isotropic Gaussian prior.
The following shows the equivalence of prior probabilities of the original and transformed weight vector for this case.
\begin{align}
	p(\bm{\theta}) = \mathcal{N}(\bm{\theta}|\bm{0}, \bm{I}) &= \frac{1}{(2\pi)^{d/2}} \exp \left(-\frac{1}{2} \bm{\theta}^\top\bm{\theta} \right) \label{formula:prior_start}\\
    &=
    \label{formula:prior_start_orth}
    \frac{1}{(2\pi)^{d/2}} \exp\left(-\frac{1}{2} \bm{\theta}^\top \bm{E}_j^\top\bm{E}_j \bm{\theta} \right)\\
	&= \frac{1}{(2\pi)^{d/2}} \exp\left(-\frac{1}{2} (\bm{E}_j\bm{\theta})^\top (\bm{E}_j \bm{\theta}) \right)\\
    &= \mathcal{N}(\bm{E}_j\bm{\theta}|\bm{0}, \bm{I}) = p(\bm{E}_j\bm{\theta}) \label{formula:prior}
\end{align}
The orthogonality of $\bm{E}_j$ follows from the fact that $\bm{E}_j = \bm{T} \bm{P}$ is the product of orthogonal matrices. $\bm{P}$ is a permutation matrix and, therefore, orthogonal. $\bm{T}$ is a diagonal matrix that is also orthogonal because the transformation it represents are designed precisely such that $\bm{T}^\top \bm{T}$ is the identity matrix (e.g., with diagonal values equal to $-1$ for tanh).
The orthogonality of $\bm{E}_j$ is used to pass from Equation~(\ref{formula:prior_start})
to Equation~(\ref{formula:prior_start_orth}).
Note, however, that scaling transformations such as those induced by ReLU can generally not be represented by orthogonal matrices.
Due to the infinite amount of possible scaling transformations, these pose a less tangible problem.

\subsubsection{Likelihood Invariance}

Similarly, we show that the likelihood is invariant to equioutput parameter states,
using the fact that $\bm{\theta}$ and $\bm{E}_j\bm{\theta}$ are equioutput parameter states:
\begin{align}
	p(\mathcal{D}| \bm{\theta}) &= \prod_{i=1}^{N} p(\bm{y}^{(i)} | \bm{x}^{(i)}, \bm{\theta}) = \prod_{i=1}^{N} p(\bm{y}^{(i)}|f_{\bm{\theta}}(\bm{x}^{(i)}))\\
    &=
    \prod_{i=1}^{N} p(\bm{y}^{(i)}|f_{\bm{E}_j\bm{\theta}}(\bm{x}^{(i)})) \\
	&= \prod_{i=1}^{N} p(\bm{y}^{(i)} | \bm{x}^{(i)}, \bm{E}_j \bm{\theta}) = p(\mathcal{D}| \bm{E}_j \bm{\theta}) \label{formula:likelihood}
\end{align}

\subsubsection{Posterior Invariance} 

Following from Equations~\ref{formula:prior_start} to \ref{formula:likelihood}, the posterior probabilities for two equioutput parameter states are identical:
\begin{align}
	p(\bm{\theta} | \mathcal{D}) &= \frac{p(\mathcal{D}|\bm{\theta}) p(\bm{\theta})}{p(\mathcal{D})}
\underset{(\ref{formula:posterior})}{\overset{(\ref{formula:likelihood})}{=}}
 \frac{p(\mathcal{D}|\bm{\bm{E}_j\theta}) p(\bm{E}_j
		\bm{\theta})}{p(\mathcal{D})} = p(\bm{E}_j\bm{\theta} | \mathcal{D}) \label{formula:posterior}
\end{align}

\subsection{Proof Sketch for Proposition~\ref{prop:param_space_dissection}} \label{proof_dissection}

Here, we provide a proof sketch for Proposition~\ref{prop:param_space_dissection}, which fundamentally builds on \cite{chen_geometry_1993} and contains some considerations on handling edge and boundary cases.
Let $\mathcal{S}_1 \subset \Theta$ be a complete set of representatives from the equivalence classes induced by the equioutput relation $\sim$, s.t. $\forall \bm{\theta} \in \Theta ~ \exists_1 \bm{\theta}^\prime \in \mathcal{S}_1: \bm{\theta} \sim \bm{\theta}^\prime$.
We call $\mathcal{S}_1$ the \textit{reference set} of uniquely identified network parameter states (cf. open minimal sufficient search sets in \cite{chen_geometry_1993}). 
Let $\mathcal{S}_j$ be the image of $\mathcal{S}_1$ under the transformation $\mathcal{F}_{\bm{E}_j}$ induced by $\bm{E}_j \in \mathcal{E}$, s.t.
$$\mathcal{S}_j = \mathcal{F}_{\bm{E}_j}(\mathcal{S}_1) =  \left \{ 
\bm{E}_j \bm{\theta}: \bm{\theta} \in \mathcal{S}_1, \bm{E}_j \in \mathcal{E} \right \}, ~~ j = 1, \dots, |\mathcal{E}|.$$

Now, we collect all edge and boundary cases in the residual set $\mathcal{S}^0$ by the following operations:
$$\mathcal{S}^0 := \Theta ~\setminus \bigcup_{j = 1}^{|\mathcal{E}|} \mathcal{S}_j$$
$$\mathcal{S}^0 = \mathcal{S}^0 \cup \bigcup_{i \neq j} (\mathcal{S}_i \cap \mathcal{S}_j)$$
\begin{align*}
    \forall j = 1, \dots, |\mathcal{E}|: {\mathcal{S}_j} &= \mathcal{S}_j \setminus \mathcal{S}^0 \\
     \Rightarrow \forall i, j \in \{ 1, \dots, |\mathcal{E}|\}, i \neq j : {\mathcal{S}_i} &\cap {\mathcal{S}_j} = \emptyset, ~ {\mathcal{S}_j} \cap \mathcal{S}^0 = \emptyset
\end{align*}
We thus have sets ${\mathcal{S}_i}, {\mathcal{S}_j}$ that are pairwise disjoint and whose intersection with the residual set $\mathcal{S}^0$ is empty.
It remains to be shown that the union over all these sets is equal to $\Theta$. 
For this, consider $\bm{\theta} \in \Theta$ with $\bm{\theta} \not \in \left( \mathcal{S}^0 \cup \bigcup_{j = 1}^{|\mathcal{E}|} {\mathcal{S}_j} \right)$.
Since the reference set contains, by definition, a representative of each equivalence class, there exists $\bm{\theta}^\prime \in \mathcal{S}_1: \bm{\theta} \sim \bm{\theta}^\prime$.
From this, however, it follows that $\exists j \in \{ 1, \dots, |\mathcal{E}|\}: \bm{\theta} = \bm{E}_j \bm{\theta}^\prime ~ \Rightarrow \bm{\theta} \in \mathcal{S}_j$, and in particular, $\bm{\theta} \in \left( \mathcal{S}^0 \cup \bigcup_{j = 1}^{|\mathcal{E}|} {\mathcal{S}_j} \right)$, such that we reach a contradiction.

\subsection{Proof for Corollary~\ref{cor:predictive}}

Combining the results from \ref{prior_inv} and \ref{proof_dissection}, it follows that the posterior predictive density
expresses as:
\begin{alignat}{3}
\label{corol_step1}
	&p(\bm{y}^\ast | \bm{x}^\ast, \mathcal{D}) &&= &&\int_{\Theta} p(\bm{y}^\ast | \bm{x}^\ast, \bm{\theta}) p(\bm{\theta} | \mathcal{D}) \diff\bm{\theta} \\
  & && = &&
  \label{corol_step2}
  \sum_{i=1}^{|\mathcal{E}|}\int_{\mathcal{S}_i} p(\bm{y}^\ast | \bm{x}^\ast, \bm{\theta}) p(\bm{\theta} | \mathcal{D}) \diff\bm{\theta} \\ \nonumber
    &&&&&+ \underbrace{\int_{\mathcal{S}^0} p(\bm{y}^\ast | \bm{x}^\ast, \bm{\theta})p(\bm{\theta}| \mathcal{D}) \diff \bm{\theta}}_{c} \\
      \label{corol_step3}
	& &&= && \sum_{i=1}^{|\mathcal{E}|}\int_{\mathcal{S}_i} p(\bm{y}^\ast | \bm{x}^\ast, \bm{\theta}) p(\bm{\theta} | \mathcal{D}) \diff\bm{\theta} + c\\
 \label{corol_step4}
 & && = &&
 \sum_{i=1}^{|\mathcal{E}|}\int_{\mathcal{S}_{1}} p(\bm{y}^\ast | \bm{x}^\ast, \bm{E}_{i}\bm{\theta}) p(\bm{E}_{i}\bm{\theta} | \mathcal{D}) \diff\bm{\theta} + c\\
	& &&= &&\int_{\mathcal{S}_1}
  \sum_{i=1}^{|\mathcal{E}|} p(\bm{y}^\ast | \bm{x}^\ast, \bm{E}_{i}\bm{\theta}) p(\bm{E}_{i}\bm{\theta} | \mathcal{D}) \diff \bm{\theta} + c\\
	& &&= &&\int_{\mathcal{S}_1} |\mathcal{E}| \cdot p(\bm{y}^\ast | \bm{x}^\ast, \bm{\theta})p(\bm{\theta}| \mathcal{D}) \diff \bm{\theta} + c\\
	& &&= &&|\mathcal{E}| \cdot \int_{\mathcal{S}_1} p(\bm{y}^\ast | \bm{x}^\ast, \bm{\theta})p(\bm{\theta}| \mathcal{D}) \diff \bm{\theta} + c \label{formula:reduced_predictive}
\end{alignat} 
Equation~(\ref{prop:param_space_dissection})
enables the transition from Equation~(\ref{corol_step1}) to~\ref{corol_step2}.
The likelihood and posterior invariance,
as stated in Equations~(\ref{formula:likelihood}) and~(\ref{formula:posterior}),
allow to rewrite Equation~(\ref{corol_step3}) as~(\ref{corol_step4}).

\subsection{Upper Bound of Markov Chains} \label{app:upper}
We now prove the upper bound derived in Equation~(\ref{formula:ccp}). We first note that Markov's inequality for the number of chains $G$ yields
    \begin{equation}
        P(G \geq \rho) \leq \mathbb{E}(G) \cdot \rho^{-1}.
    \end{equation}
We can rewrite this inequality as 
    \begin{equation} \label{eq:upb1}
        P(G < \rho) \geq 1- \mathbb{E}[G] \cdot \rho^{-1}.
    \end{equation}

The number of independent chains that are required to visit every mode at least once can be framed as a generalized coupon collector's problem\cite{flajolet_birthday_1992}, where the number of draws (independent chains) needed to collect (visit) all $\nu$ coupons (modes) is estimated. This gives us an expression for the expected number of chains:
\begin{align} \label{eq:upb2}
    \mathbb{E}(G) &= \sum_{q = 0}^{\nu - 1}(-1)^{\nu - 1 - q} \sum_{J: |J| = q} (1 - \Pi_J)^{-1}, \quad \text{with } \Pi_J = \sum_{j \in J} \pi_j.
\end{align}
Putting together Equation~(\ref{eq:upb1}) and Equation~(\ref{eq:upb2}), we get Equation~(\ref{formula:ccp}).

\section{Posterior Symmetry Removal} \label{app:symmetry_removal}
Section \ref{subsec:upperbound} uses the knowledge of equioutput parameter states to derive an upper bound on the number of Markov chains required to observe all functionally diverse modes of the MLP parameter posterior density.
However, the obtained samples are scattered across the sets $\mathcal{S}_j$; see Section~\ref{sec:efficient_sampling} and Equation~(\ref{formula:union}).
This obscures any insights into the geometry of the network's parameter posterior density and makes it infeasible to interpret anything but the obtained functional uncertainty.
Ideally, all samples should therefore be mapped to a representative set.
In the following, we derive a symmetry removal algorithm for tanh-activated MLPs.

\subsubsection{Reasoning.}
Assume $G$ available posterior samples $\bm{\theta}^{(g)}, g \in \{1, \dots, G\}$.
As equioutput transformations in MLPs factorize layerwise (see Section~\ref{sec:background}), symmetries can be removed by iterating through the $K - 2$ hidden network layers.
We propose to operate on the layers in reverse order, motivated by the idea that the output of the previous layer $l-1$ can be interpreted as an input to a single-layer MLP with $M_{l}$ neurons.
Therefore, processing an MLP in this way is comparable to removing symmetries from $K-2$ independent single-layer MLPs.

In each step, it is sufficient to consider the parameters of neurons from the current hidden layer $l$.
For a hidden neuron $i \in \{1, \dots, M_l\}$ in the $l$-th layer, the corresponding parameters are the weights and biases from the neuron output $o_{li}$ and weights connecting the neuron to the following layer.
The neuron parameter vector
$\bm{\phi}^{(g, l, i)}\in \mathbb{R}^{(M_{(l-1)} + M_{(l+1)} + 1)}$
is defined by
\[\bm{\phi}^{(g, l, i)} = (w^{(g)}_{li1}, \dots, w^{(g)}_{liM_{(l - 1)}}, w^{(g)}_{(l+1)1i}, \dots, w^{(g)}_{(l+1) M_{(l+1)}i}, b^{(g)}_{li})^\top,\]
with $w^{(g)}_{lij}, b^{(g)}_{li}$ for $2 \leq l \leq K - 1, 1 \leq i \leq M_l, 1 \leq j \leq M_{(l-1)}$.
This vector is an element of a subspace of $\Theta$ whose dimensionality depends on the width of the previous and subsequent layer.
As neurons can be freely interchanged (see Section~\ref{sec:background}), the marginal posteriors of freely permutable neurons from the same hidden layer are identical.
This implies that the marginal density of every $\bm{\phi}^{(g, l, i)}$ contains at least $M_l$ different regions to which a neuron might be assigned in its mapping to the reference set, allowing for $M_l!$ different permutation configurations.
The permutation-related symmetries can be removed by finding a valid reference assignment of states to hidden neurons, which effectively dissects the parameter subspace into $M_l$ regions.
For tanh-activated MLPs, as considered in the subsequent paragraph, the number of states is further doubled due to sign-flip equioutput parameter states.
Therefore, prior to the permutation symmetry removal, we cut the parameter subspace in half in a geometry-respecting manner for tanh-related symmetry removal (details below).
In the following, we present an exemplary algorithm for automatic symmetry removal in tanh-activated MLPs without the need to explicitly specify $\mathcal{E}$.

\subsubsection{Detailed Algorithm.}
We need to find identifiability constraints, i.e., a selection of hyperplanes in parameter space that distribute the available space uniquely to the neurons of a layer.
As a result, neurons cannot possess symmetric configurations, which effectively removes symmetries.
For this, we follow the principle from \citet{fruhwirth-schnatter_markov_2001} and work layer-wise in the subspace of the neuron parameters that are involved in the corresponding equioutput transformations of that layer.
Let $\mathcal{M}^{(g, l)} := \{\bm{\phi}^{(g, l, 1)}, \dots, \bm{\phi}^{(g, l, M_l)}\}$ be the collection of all parameters of the $l$-th layer for a sample $g$, and let $\mathcal{M}^{(*, l)} =  \bigcup_{g=1}^G \mathcal{M}^{(g, l)}$.
In order to remove the tanh-related symmetries from a layer $l$, the parameter subspace of the respective neurons must be reduced by half.
Halving the space while respecting the geometry of the posterior density poses a data-dependent optimization problem of finding the right constraints. Here, we propose a zero-centered linear hyperplane $\bm{\beta}_l \in \mathbb{R}^{M_{(l-1)} + M_{(l+1)} + 1}$ that intersects as few states or clusters of neuron parameter vectors $\bm{\phi} \in \mathcal{M}^{(*, l)}$ as possible.
We optimize this hyperplane using a variant of a support vector machine (SVM; \cite{cortes_support-vector_1995}) for unlabeled data, such that it has a maximum distance to all neuron parameter vectors. 
The loss function follows the hinge-loss formulation of SVMs but substitutes the labels for absolute values of $\bm{\beta}_l^\top\bm{\phi}$ for $\bm{\phi} \in \mathcal{M}^{(*, l)}$, taking the form
\begin{equation*}
\mathcal{L}(\bm{\beta}_l) = \tfrac{1}{2} \bm{\beta}_l^\top\bm{\beta}_l + C \cdot \textstyle\sum_{\bm{\phi}\in\mathcal{M}^{(*, l)}}\max \left(0, 1 - \left |\bm{\beta}_l^\top\bm{\phi} \right| \right),
\end{equation*}
where $C>0$ is a cost-related hyperparameter and $|\cdot|$ a user-defined norm (we use the $L_1$ norm in analogy to the hinge loss).
The loss-minimal hyperplane is chosen as a geometry-respecting constraint and is used for flipping parameter vectors to one side of the hyperplane, i.e., applying the a tanh-related equioutput transformation and flipping the signs of the corresponding parameter vector entries (Algorithm~\ref{algo:tanh_removal}).
In practice we 
run $K_{\bm{\beta}} > 1$ optimizations with randomly initialized hyperplanes since the global optimum is not guaranteed from this SVM variation.

\begin{algorithm}[H]
	\caption{
        Geometry-respecting tanh removal algorithm.
    }
	\label{algo:tanh_removal}
	\begin{algorithmic}
		\Procedure{TanhRemoval}{$\mathcal{M}^{(*, l)}$, $K_{\bm{\beta}}$}
		\State $\mathcal{R} \gets \emptyset$
		\For{each $k$ in range $K_{\bm{\beta}}$}
		\State $\bm{\beta}_k \gets \argmin_{\bm{\beta}} \mathcal{L}(\bm{\beta})$
		\State $\mathcal{R} \gets \mathcal{R} \cup \{\bm{\beta}_k\}$
		\EndFor
		\State $\bm{\beta}^\ast \gets \bm{\beta} \in \mathcal{R}$ with minimal loss $\mathcal{L}(\bm{\beta})$
        \For{each $\bm{\phi} \in \mathcal{M}^{(*, l)}$}
            \If{$\bm{\beta}^{*\top}\bm{\phi} < 0$}
                \State $\bm{\phi}$ $\gets$ $-\bm{\phi}$
            \EndIf
        \EndFor
		\EndProcedure
	\end{algorithmic}
\end{algorithm}

Following the removal of tanh-related symmetries using the SVM approach, the remaining parameter subspace in the marginal posterior density of the parameters in layer $l$ must be divided among the hidden-layer neurons in order to remove the permutation-related symmetries.
For this, we consider one sample $\bm{\theta}^{(g)}$ with elements $\bm{\phi}^{(g, l, i)} \in \mathcal{M}^{(g, l)}$ at a time and assign classification labels $c^{(g, l, i)}, i \in \{1$, $\dots, M_l\}$.
Initially, each element is labeled with its neuron index $i$.
We then perform a greedy constrained $k$-NN classification (Algorithm~\ref{algo:gjknnc}) on the elements of each set $\mathcal{M}^{(g, l)}$ based on their pairwise spatial distances using the Gaussian similarity function $s(a, b) = \exp\left(-||a - b||^2 \cdot (2\sigma^2)^{-1}\right)$ with $\sigma = 1$ to assign each neuron parameter vector $\bm{\phi}^{(g, l, i)}$ to its most likely neuron position in the hidden layer.
The hyperparameter $k$ should be carefully selected. We generally recommend high values, such that the classification is based on as many neighbors as possible. We set $k=1024$ in all experiments.

\begin{algorithm}[H]
	\caption{
        Algorithm to perform greedy constrained $k$-NN classification.
    }
	\label{algo:gjknnc}
	\begin{algorithmic}
		\Procedure{GreedyConstrainedKNNClassification}{$\mathcal{M}^{(g, l)}$, $\mathcal{M}^{(*, l)}$, $k$}
		\State $\mathcal{R} \gets \emptyset$
		\For{each $i$ in range $M_l$}
		\State $\phi^{(g,l,i)} \in \mathcal{M}^{(g, l)}$
		\State $(p_i(c=1), \dots, p_i(c=M_l)) \gets $ \textsc{KNNClassification}($\bm{\phi}^{(g, l, i)}$, $\mathcal{M}^{(*, l)}$, $k$)
		\State $\mathcal{R} \gets \mathcal{R} \cup \{(p_i(c=1), \dots, p_i(c=M_l))\}$
		\EndFor
        \State $i \gets 1$
        \Repeat
		\State Find the prob. vector $p_j$ from $\mathcal{R}$ with highest probability for a class $c^\ast$.
		\State $c^{(g, l, j)} \gets \argmax_c p_j(c=c^\ast)$
        \State $i \gets i + 1$
        \Until{$i = M_l$}
		\State Set $p_j(c=c^\ast) = 0$ for all $j$.
		\State \Return $\{c^{(g, l, 1)}, \dots, c^{(g, l, M_l)}\}$
		\EndProcedure
	\end{algorithmic}
\end{algorithm}

This is followed by a permutation of the neuron parameter vectors of the layer according to the classification results, i.e., neurons are re-indexed to their class assignment (Algorithm~\ref{algo:permutation_removal}) in the \textsc{permute} operation.
This effectively clusters the parameter subspace into $M_l$ regions that implicitly define constraints and remove permutation-related symmetries.
The $k$-NN classification is constrained in such a way that no pair of neuron parameter vectors $(\bm{\phi}^{(g, l, i)}, \bm{\phi}^{(g, l, j)})$ in the set $\mathcal{M}^{(g, l)}$ is allowed to have the same class.
This is done greedily by assigning the element with the highest probability for any class first, followed by the remaining elements in decreasing order of probabilities, until all vectors are assigned.
The process is repeated for $I$ iterations over all $K-1$ hidden layers and $G$ Monte Carlo samples to obtain a globally coherent clustering of neuron parameter vectors, since the classification by the greedy constrained $k$-NN classification algorithm is based on local relationships.
We set $I = 256$ in all experiments.

\begin{algorithm}[H]
	\caption{
        Permutation symmetry removal for a hidden layer $l$.
    }
	\label{algo:permutation_removal}
	\begin{algorithmic}
		\Procedure{PermutationRemoval}{$\mathcal{M}^{(1, l)}$, $\dots$, $\mathcal{M}^{(G, l)}$, $\mathcal{M}^{(*, l)}$, $k$, $I$}
		\For{$I$ times}
		\For{each $g$ in range $G$}
		\State $\{c^{(g, l, 1)}, \dots, c^{(g, l, M_l)}\} \gets $ \\ \phantom{procedure for~~} \textsc{GreedyConstrainedKNNClassification}($\mathcal{M}^{(g, l)}$, $\mathcal{M}^{(*, l)}$, $k$)
		\EndFor
		\For{each $s$ in range $S$}
		\State \textsc{permute}($\mathcal{M}^{(g, l)}$, $\{c^{(g, l, 1)}, \dots, c^{(g, l, M_l)}\}$)
		\EndFor
		\EndFor
		\EndProcedure
	\end{algorithmic}
\end{algorithm}

Algorithm~\ref{algo:complete_removal} outlines
the entire algorithm for removing symmetries
from a Markov chain sampled from
an MLP parameter posterior density.
Algorithm~\ref{algo:complete_removal}
iteratively invokes
Algorithms~\ref{algo:tanh_removal}
and~\ref{algo:permutation_removal}.

\begin{algorithm}[H]
	\caption{Complete geometry-respecting symmetry removal algorithm.}
	\label{algo:complete_removal}
	\begin{algorithmic}
		\Procedure{GeometryRemoval}{$\{\bm{\theta}^1, \dots, \bm{\theta}^G\}$, $I$, $k$, $K_{\bm{\beta}}$}
		\For{each layer $l$ (reverse)}
		\State Construct neuron parameter sets $M^{(1, l)}, \dots, M^{(G, l)}$ from MCMC samples $\{\bm{\theta}^1, \dots, \bm{\theta}^G\}$
		\State $\mathcal{M}^{(*, l)} \gets \mathcal{M}^{(1, l)} \cup \dots \cup \mathcal{M}^{(G, l)}$
		\State \textsc{TanHRemoval}($\mathcal{M}^{(*, l)}$, $K_{\bm{\beta}}$); see Algorithm~\ref{algo:tanh_removal}
		\State \textsc{PermutationRemoval}($\mathcal{M}^{(1, l)}$, $\dots$, $\mathcal{M}^{(G, l)}$, $\mathcal{M}^{(*, l)}$, $k$, $I$); see Algorithm~\ref{algo:permutation_removal}
		\EndFor
		\EndProcedure
	\end{algorithmic}
\end{algorithm}

\section{Experimental Setup} \label{app:experimental_setup}
\subsection{Datasets} \label{app:datasets}
\begin{figure}[H]
	\includegraphics[width=1.0\linewidth]{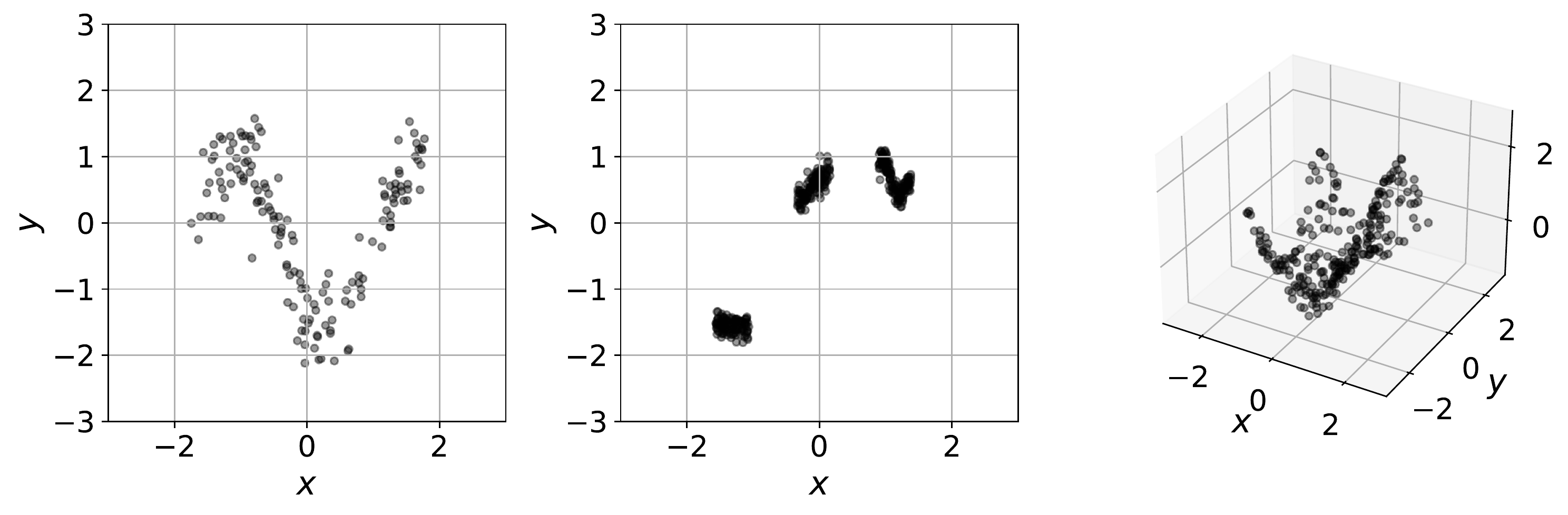}
	\caption{Visualization of synthetic datasets $\mathcal{D}_S$ (left), $\mathcal{D}_I$ (center), and $\mathcal{D}_R$ (right).}
	\label{fig:datasets}
\end{figure}

The sinusoidal dataset $\mathcal{D}_S$ is adopted from an example provided in \citet{daxberger_2021_LaplaceReduxEffortless}, and the dataset $\mathcal{D}_I$ is a synthetic dataset from \citet{izmailov_subspace_2020}.
The Regression2d dataset $\mathcal{D}_R$ is another synthetic dataset that we generate as follows:
\begin{align*}
	p(x_1) &= \mathcal{U}(a=-2.0, b=2.0) \\
	p(x_2) &= \mathcal{U}(a=-2.0, b=2.0) \\
	f(x_1, x_2) &= x_1 \cdot \sin(x_1) + \cos(x_2) \\
	p(y | x_1, x_2) &= \mathcal{N}(\mu = f(x_1, x_2), \sigma = 0.1) \\
	p(\mathcal{D}) &=
 \prod_{i=1}^{256} p(y^{(i)}| x_1^{(i)}, x_2^{(i)}) p(x_1^{(i)})p(x_2^{(i)})\label{formula:regression2d}
\end{align*}
The three synthetic datasets
$\mathcal{D}_S,\mathcal{D}_I$ and $\mathcal{D}_R$ were standardized for all experiments and are visualized in Figure~\ref{fig:datasets}.
We split the data in 80\% training and 20\% test observations.
Together with the synthetic datasets we further provide descriptions and references also for the UCI datasets used in our benchmarks in Table~\ref{tab:further}.

\begin{table}[H]
\begin{center}
\caption{Dataset characteristics and references.}
\vspace{0.3cm}
\label{tab:further}
\begin{footnotesize}
\begin{tabular}{lrrrrrll}
Dataset & Total Observations & \# Train & \# Test & Features && Reference \\ \hline 
Sinusoidal ($\mathcal{D}_S$) & 150 & 120 & 30 & 1 && Adapted from \cite{daxberger_2021_LaplaceReduxEffortless} \\
Izmailov ($\mathcal{D}_i$) & 400 & 320 & 80 & 1 && Adapted from \citet{izmailov_subspace_2020} \\
Regression2d ($\mathcal{D}_R$) & 256 & 304 & 52 & 2 && -- \\
Airfoil & 1503 & 1202 & 301 & 5 && \citet{Dua.2019}  \\
Concrete & 1030 & 824 & 206 & 8 &&  \citet{yeh1998modeling} \\
Diabetes & 442 & 353 & 89 & 10  &&  \citet{Dua.2019} \\
Energy & 768 & 614 & 154 & 8  &&  \citet{tsanas2012accurate} \\
Forest Fire & 517 & 413 & 104 & 12 &&  \citet{cortez2007data} \\
Yacht & 308 & 246 & 62 & 6 &&  \citet{Dua.2019}\\ \hline
\end{tabular}
\end{footnotesize}
\end{center}
\end{table}

\subsection{Markov Chain Monte Carlo} \label{app:mcmc}
MCMC sampling of the MLP parameters is performed using the \texttt{NumPyro} implementation of the No U-Turn Sampler (NUTS) with default settings.
Specifically, we only provide our models and data to NUTS, set the step size to $1.0$ and allow for adaptation during the warm-up phase.
We use a diagonal inverse mass matrix which is initialized with the identity matrix and is allowed to adapt during the warm-up phase.
The target acceptance probability is set to $0.8$.

\subsection{Point Estimates} \label{app:point_estimates}
For the LA and DE estimates, both the smaller architecture $f_1$ (one hidden layer of three neurons) and the larger architecture $f_2$ (three hidden layers of 16 neurons each) are trained with an RMSProp optimizer \cite{hinton_2012_RMSPropDivideGradient} for 500 and 1000 epochs, respectively, using a constant learning rate of $10^{-4}$ and no weight decay.
In the case of DE, we initialize each weight vector with a different random seed and do not use data bootstrapping, following the approach described in the original work of \citet{lakshminarayanan_simple_2017}.
LA samples are drawn directly from the parameter posterior density.
Due to the small dataset sizes, we perform full-batch training.
Our loss function is set to be the negative log-posterior
\begin{align}
    \mathcal{L}(\bm{\theta}, \sigma) = - \log p(\bm{\theta} | \mathcal{D})
    = \frac{1}{2\sigma^2} \textstyle\sum_{i=1}^N (f_{\bm{\theta}}(\bm{x}_i) - \bm{y}_i)^2 + N \cdot \log\sigma + \tfrac{1}{2} \bm{\theta}^\top \bm{\theta}.
\end{align}
The loss function is to be minimized over the network parameters $\bm{\theta}$ and nuisance parameter $\sigma$.
The code for LA and DE estimates is mainly based on the \texttt{PyTorch} \cite{Paszke2019} and \texttt{PyTorch Lightning} \cite{Falcon2019} libraries, as well as the \texttt{Laplace} library by \citet{daxberger_2021_LaplaceReduxEffortless}.

\section{Methods} \label{app:methods}
\subsection{KL-Divergence for the Posterior Predictive Density} \label{app:kl_divergence}
The KL-divergence for two densities $p, q$ of continuous random variables is defined as
\begin{align}
	D_{KL} (p || q) &= \int_{\mathcal{X}} p(x) \log \frac{p(x)}{q(x)} \diff x.
\end{align}
The analogous formulation for random variables with discrete probability densities  $p, q$ is 
\begin{align}
	D_{KL} (p || q) &= \sum_{x\in \mathcal{X}} p(x) \log \frac{p(x)}{q(x)}.
\end{align}
In Section~\ref{subsec:results_corollary} the KL-divergence of consecutive posterior predictive densities is calculated on a regular grid of values $y \in [-3.0, 3.0]$ for a particular input $x$.
Further, this KL-divergence is averaged over a regular grid of input values $x \in [-3.0, 3.0]$ to obtain the final value.

\subsection{Spectral Clustering} \label{app:spectral_clustering}
The spectral clustering in Section~\ref{subsec:results_symmetry_removal} is performed by first constructing a 4-NN graph using the Gaussian similarity function $s(a, b) = \exp\left(-||a - b||^2 \cdot (2\sigma^2)^{-1}\right)$, with $\sigma = 1$ as a distance measure, encoded via the adjacency matrix $\bm{A}$.
We then compute the normalized graph Laplacian $\bm{L}_{norm}$ from $\bm{A}$ as
\begin{align}
	\bm{D}_{ij} &= \begin{cases}
		\sum_{k=1}^{N} \bm{A}_{ik} & \mbox{if } i = j \\
		0 & \mbox{else} \\
	\end{cases}, \\
	\bm{L}_{norm} &= \bm{D}^{-1/2} \bm{L} \bm{D}^{1/2},
\end{align}
where $\bm{D}_{ij}$ is the $(i, j)$ element of the degree matrix
$\bm{D}$ of the graph.
Subsequently, the eigenvalue spectrum of $\bm{L}_{norm}$ is determined and K-means clustering is performed for $K=3$.

\section{Additional Results} \label{app:results_additional}
\subsection{Performance Comparison Between our Approach and LA} \label{app:results_performance}
We compare our approach of calculating the PPD based on MCMC sampling to LA in terms of predictive performance.
For our MCMC sampling, we generate 1274 chains, which is the derived upper bound for the number of chains.
Table~\ref{tab:result_posteriors_la} summarizes this comparison.

\begin{table}[ht]
    \caption{Mean log pointwise predictive density (LPPD) values on test sets (larger is better; one standard error in parentheses). The highest performance per dataset and network is highlighted in bold.}
    \resizebox{1.0\textwidth}{!}{
    \begin{tabular}{l|rrr|rrr}
    \multicolumn{1}{c}{} & \multicolumn{3}{c}{\textbf{Smaller network $f_1$}} & \multicolumn{3}{c}{\textbf{Larger network $f_2$}} \\
    \multicolumn{1}{c}{} & \multicolumn{1}{c}{MCMC (ours)} & \multicolumn{1}{c}{MCMC (s.c.)} & \multicolumn{1}{c}{LA} & \multicolumn{1}{c}{MCMC (Ours)} & \multicolumn{1}{c}{MCMC (s. c.)} & \multicolumn{1}{c}{LA} \\ \hline
    $\mathcal{D}_S$ & \textbf{-0.53} ($\pm$ 0.09) & -0.56 ($\pm$ 0.11) & -0.57 ($\pm$ 0.10) & \textbf{-0.59} ($\pm$ 0.12) & \textbf{-0.59} ($\pm$ 0.12) & -2.42 ($\pm$ 0.01) \\
    $\mathcal{D}_I$ & \textbf{0.79} ($\pm$ 0.06) & 0.65 ($\pm$ 0.07) & 0.53 ($\pm$ 0.07) & \textbf{0.91} ($\pm$ 0.09) & \textbf{0.91} ($\pm$ 0.09) & -1.81 ($\pm$ 0.01) \\  
    $\mathcal{D}_R$ & 0.64 ($\pm$ 0.10) & \textbf{0.75} ($\pm$ 0.11) & -27.39 ($\pm$ 3.65) & \textbf{0.95} ($\pm$ 0.08) & \textbf{0.95} ($\pm$ 0.08) & -2.33 ($\pm$ 0.00) \\ 
    Airfoil & \textbf{-0.74} ($\pm$ 0.04) & -0.80 ($\pm$ 0.05) & -1.78 ($\pm$ 0.13) & \textbf{0.92}  ($\pm$ 0.05) & 0.72 ($\pm$ 0.10) & -3.57 ($\pm$ 0.18) \\ 
    Concrete & \textbf{-0.41} ($\pm$ 0.05) & -0.44 ($\pm$ 0.06) & -14.49 ($\pm$ 1.02) & \textbf{0.26} ($\pm$ 0.07) & 0.25 ($\pm$ 0.07) & -4.36 ($\pm$ 0.47) \\ 
    Diabetes & \textbf{-1.20} ($\pm$ 0.07) & \textbf{-1.20} ($\pm$ 0.07) & -1.46 ($\pm$ 0.09) & \textbf{-1.18} ($\pm$ 0.08) & -1.22 ($\pm$ 0.09) & -2.61 ($\pm$ 0.00) \\ 
    Energy & \textbf{0.92} ($\pm$ 0.04) & 0.69 ($\pm$ 0.12) & -31.74 ($\pm$ 1.88) & 2.07 ($\pm$ 0.46) & \textbf{2.38} ($\pm$ 0.11) & -1.39 ($\pm$ 0.06) \\
    ForestF & \textbf{-1.37} ($\pm$ 0.07) & \textbf{-1.37} ($\pm$ 0.07) & -2.39 ($\pm$ 0.16) & \textbf{-1.43} ($\pm$ 0.45) & -1.69 ($\pm$ 0.49) & -2.80 ($\pm$ 0.00) \\ 
    Yacht & \textbf{1.90} ($\pm$ 0.16) & 1.29 ($\pm$ 0.56) & -5.60 ($\pm$ 1.51) & \textbf{3.31} ($\pm$ 0.21) & 0.15 ($\pm$ 0.09) & -2.69 ($\pm$ 0.00) 
    \end{tabular}
    }
    \label{tab:result_posteriors_la}
\end{table}


\subsection{Interpretability Example Approximated} \label{app:results_application_example}
\begin{figure}[H]
	\includegraphics[width=1.0\textwidth]{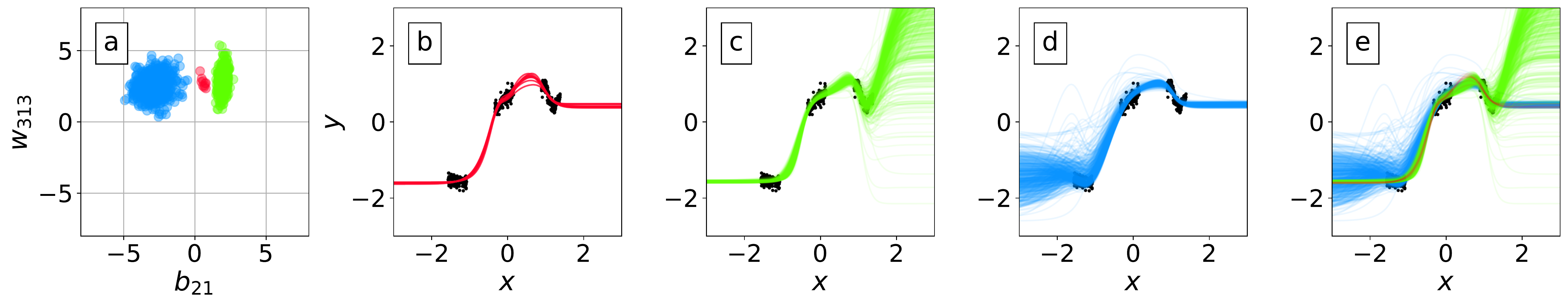}
	\caption{From left to right: The three remaining modes in the BNN's posterior after clustering approximated by MoLA, visualized in the bivariate marginal space of two weights (a);  resulting functionally diverse network parameter states based on the three Gaussian components (b-d);  the resulting function space as a composition of the samples of the three Gaussian distributions (e) by combining b-d.}
	\label{fig:mola}
\end{figure}

The three remaining modes in the BNN's posterior of the interpretability example in Section~\ref{subsec:results_symmetry_removal} can be analytically approximated using a mixture of LA (MoLA; \cite{eschenhagen_2021_MixturesLaplaceApproximations}) upon clustering.
Figure~\ref{fig:mola} depicts the approximated function space component-wise and as a mixture.

\subsection{Profile of Parameter Posterior Density} \label{app:results_full_posterior}
\begin{figure}[H]
	\includegraphics[width=1.0\textwidth]{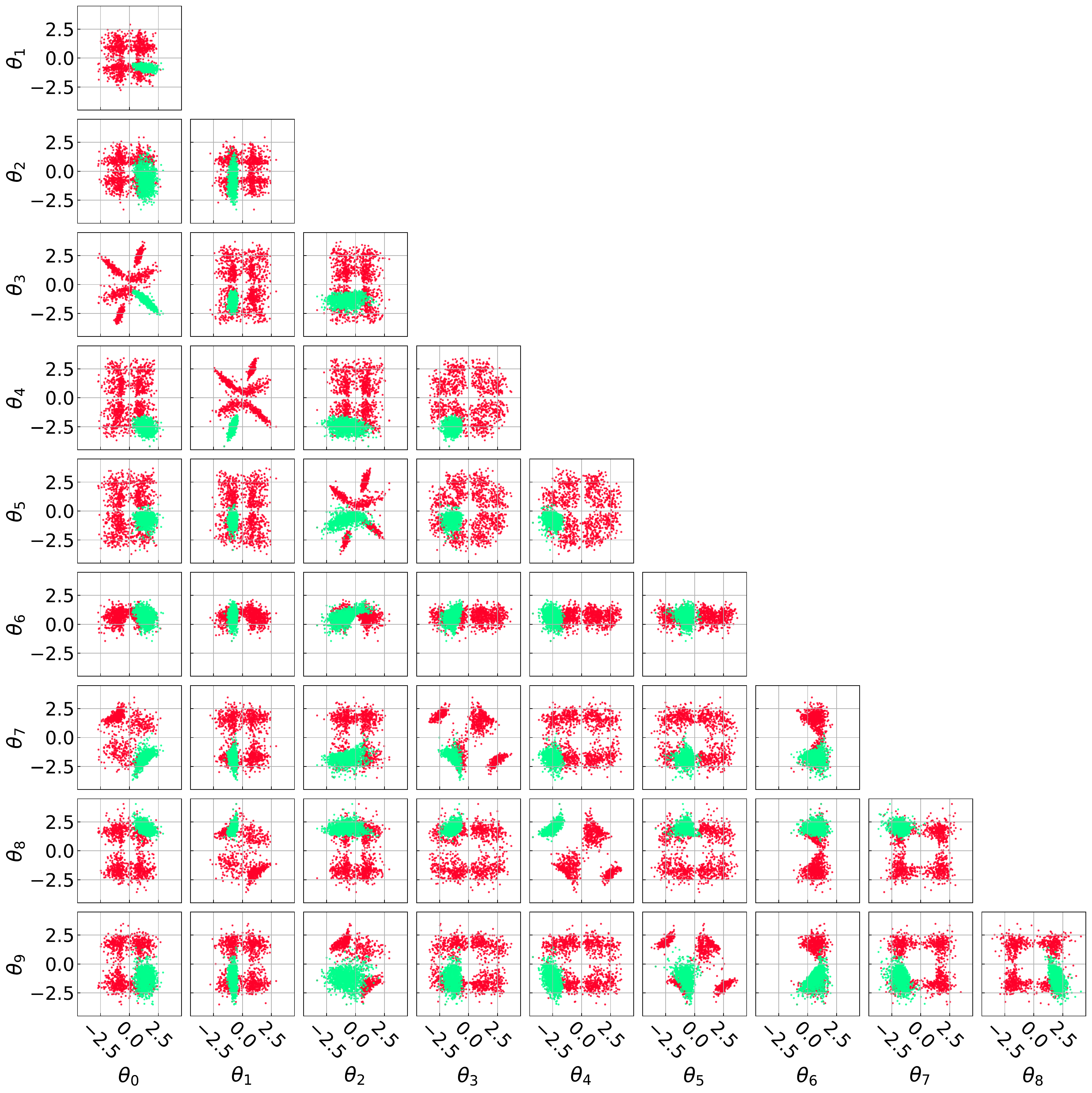}
	\caption{Visualization of pairwise profiles of the parameter posterior density of BNN $f_1$.
 The parameter posterior density sampled via MCMC (red) exhibits symmetries, while the transformed approximate parameter posterior density (green) is unimodal.}
	\label{fig:full_posterior}
\end{figure}

Figure~\ref{fig:full_posterior} visualizes
pairwise profiles of
the parameter posterior density of BNN $f_1$ on dataset $\mathcal{D}_S$ as investigated in Section~\ref{subsec:results_symmetry_removal} before (red) and after (green) the application of the symmetry removal algorithm.
The resulting approximate parameter posterior density is unimodal and much simpler
in comparison to the original parameter posterior density, yet, functionally, they are identical.

\end{document}